\documentclass[journal]{IEEEtran}
\usepackage{amsmath,amsfonts}
\usepackage{amssymb}
\newcommand{\Rmnum}[1]{\uppercase\expandafter{\romannumeral #1}}
\newcommand{\rmnum}[1]{\romannumeral #1} 
\usepackage{makecell}
\usepackage{algorithmic}
\usepackage{array}
\usepackage[caption=false,font=normalsize,labelfont=sf,textfont=sf]{subfig}
\newtheorem{remark}{Remark}
\usepackage{textcomp}
\usepackage{stfloats}
\usepackage{url}
\usepackage{verbatim}

\usepackage{graphicx}
\usepackage{hyperref}  
\usepackage{booktabs}
\usepackage{multirow}
\usepackage{colortbl} 
\usepackage[numbers,sort&compress]{natbib}
\usepackage[table,xcdraw]{xcolor}
\def\BibTeX{{\rm B\kern-.05em{\sc i\kern-.025em b}\kern-.08em
    T\kern-.1667em\lower.7ex\hbox{E}\kern-.125emX}}
\usepackage{balance}
\begin{document}
\title{\huge From Limited Labels to Open Domains: An Efficient Learning Method for Drone-view Geo-Localization} 
\author{
\thanks{}}
\author{Zhongwei Chen, \textit{Student Member, IEEE}, Zhao-Xu Yang, \textit{Member, IEEE}, Hai-Jun Rong,  \textit{Senior Member, IEEE}, \\Jiawei Lang, Guoqi Li, \textit{Member, IEEE}\\
\thanks{This paper is submitted for review on March 6, 2025. This work was supported in part by the Key Research and Development Program of Shaanxi, PR China (No. 2023-YBGY-235), the National Natural Science Foundation of China (No. 61976172 and No. 12002254), Major Scientific and Technological Innovation Project of Xianyang, PR China (No. L2023-ZDKJ-JSGG-GY-018), and Shanghai NeuHelium Neuromorphic Technology Co., Ltd.  (Corresponding author: Zhao-Xu Yang and Hai-Jun Rong)}

\thanks{Zhongwei Chen, Zhao-Xu Yang, Hai-Jun Rong and Jiawei Lang are with the State Key Laboratory for Strength and Vibration of Mechanical Structures, Shaanxi Key Laboratory of Environment and Control for Flight Vehicle, School of Aerospace Engineering, Xi’an Jiaotong University, Xi’an 710049, PR China (e-mail:ISChenawei@stu.xjtu.edu.cn; yangzhx@xjtu.edu.cn; hjrong@mail.xjtu.edu.cn). 
{Code is available at {\href{https://github.com/ISChenawei/CDIKTNet}{\color{red}{here}}}}.

Guoqi Li is with the Institute of Automation, Chinese Academy of Sciences, Beijing 100190, China, also with the School of Artificial Intelligence, University of Chinese Academy of Sciences, Beijing 101408, China, and
also with the Peng Cheng Laboratory, Shenzhen 518000, China (e-mail: guoqi.li@ia.ac.cn).}}

\maketitle

\begin{abstract}
Traditional supervised drone-view geo-localization (DVGL) methods heavily depend on paired training data and encounter difficulties in learning cross-view correlations from unpaired data. Moreover, when deployed in a new domain, these methods require obtaining the new paired data and subsequent retraining for model adaptation, which significantly increases computational overhead. Existing unsupervised methods have enabled to generate pseudo-labels based on cross-view similarity to infer the pairing relationships. However, geographical similarity and spatial continuity often cause visually analogous features at different geographical locations. The feature confusion compromises the reliability of pseudo-label generation, where incorrect pseudo-labels often drive negative optimization. Given these challenges inherent in both supervised and unsupervised DVGL methods, we propose a novel cross-domain invariant knowledge transfer network (CDIKTNet) with limited supervision, whose architecture consists of a cross-domain invariance sub-network (CDIS) and a cross-domain transfer sub-network (CDTS). This architecture facilitates a closed-loop framework for invariance feature learning and knowledge transfer. The CDIS is designed to learn cross-view structural and spatial invariance from a small amount of paired data that serves as prior knowledge. It endows the shared feature space of unpaired data with implicit cross-view correlations at initialization, which alleviates feature confusion. Based on this, CDTS employs dual-path contrastive learning to further optimize each subspace while preserving consistency in a shared feature space. Extensive experiments demonstrate that CDIKTNet achieves highly competitive performance under full supervision among existing supervised methods, and further surpasses existing unsupervised methods in both few-shot initialization and cross-domain initialization settings.
\end{abstract}
\begin{IEEEkeywords}
drone-view geo-localization, structural invariance learning, spatial invariance learning, knowledge transfer.
\end{IEEEkeywords}
\section{Introduction}

\IEEEPARstart{C}{ross-view} geo-localization has emerged as a promising visual localization technique \cite{shore2024bev}. Early efforts primarily focus on retrieving ground-view images to geo-referenced satellite-view images for ground-view geo-localization \cite{Cai_2019_ICCV}. With the widespread adoption of drones in complex scenarios such as emergency response \cite{Liu_Multi} and urban surveys \cite{Li_PVT++}, the demand for autonomous localization in GPS-denied environments has been steadily increasing. This has driven more research interests in drone-view geo-localization (DVGL) \cite{deuser2023sample4geo,wang2024multiple}. However, the dynamic viewpoints of drones result in scale distortions and non-rigid spatial transformations compared with satellite-view images. This pronounced structural and spatial heterogeneity presents challenges for DVGL. In recent years, benefiting from previous research efforts \cite{wu2024camp,chen2024multi}, supervised DVGL methods with paired data have made notable progress. These paired data often require manual labeling, which provides an accurate pairing relationship between satellite-view and drone-view images of the same location.

\label{sec:intro}

\begin{figure}[t]
  \centering
  \includegraphics[width=3.4in]{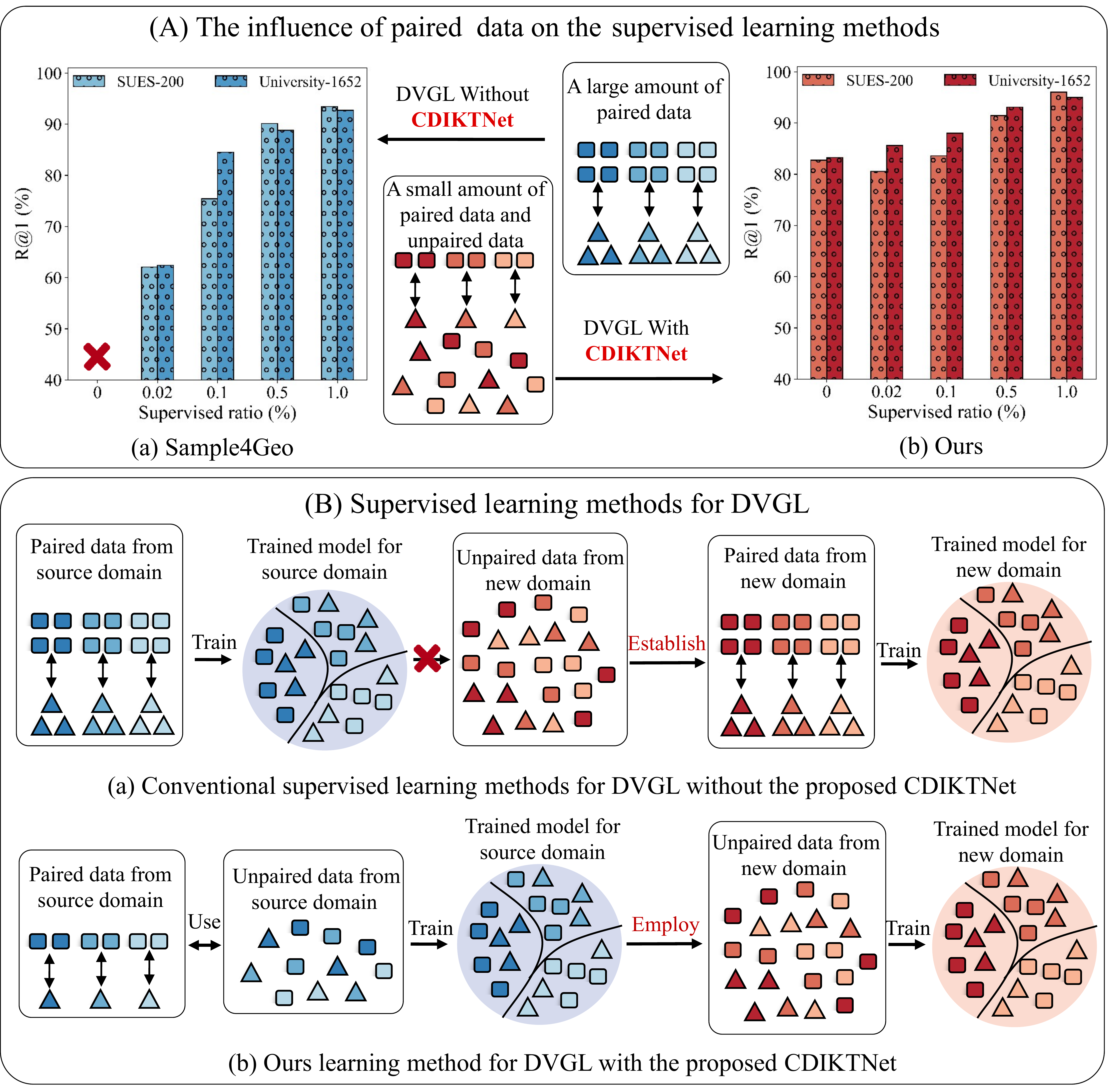}
  \caption{\textbf{CDIKTNet VS. Existing Supervised Methods}. (A) The performance variations of the representative method Sample4Geo\cite{deuser2023sample4geo} and CDIKTNet under different proportions of paired data are analyzed in the drone$\rightarrow$satellite scenario using University-1652\cite{zheng2020university} and SUES-200\cite{zhu2023sues} datasets. The evaluation on SUES-200 is performed at an altitude of 150m. It is evident that Sample4Geo generally underperforms CDIKTNet. Specifically, at a 0\% supervision ratio, Sample4Geo becomes completely ineffective, whereas CDIKTNet continues to show its effectiveness. (B) Existing supervised methods are at risk of failure when transferred to a new domain if paired data is unavailable, as they heavily rely on such data for adaptation. In contrast, CDIKTNet can adapt to new domains without requiring explicit pairing.}
  \label{fig1}
\end{figure}

\begin{figure}[t]
  \centering
  \includegraphics[width=3.4in]{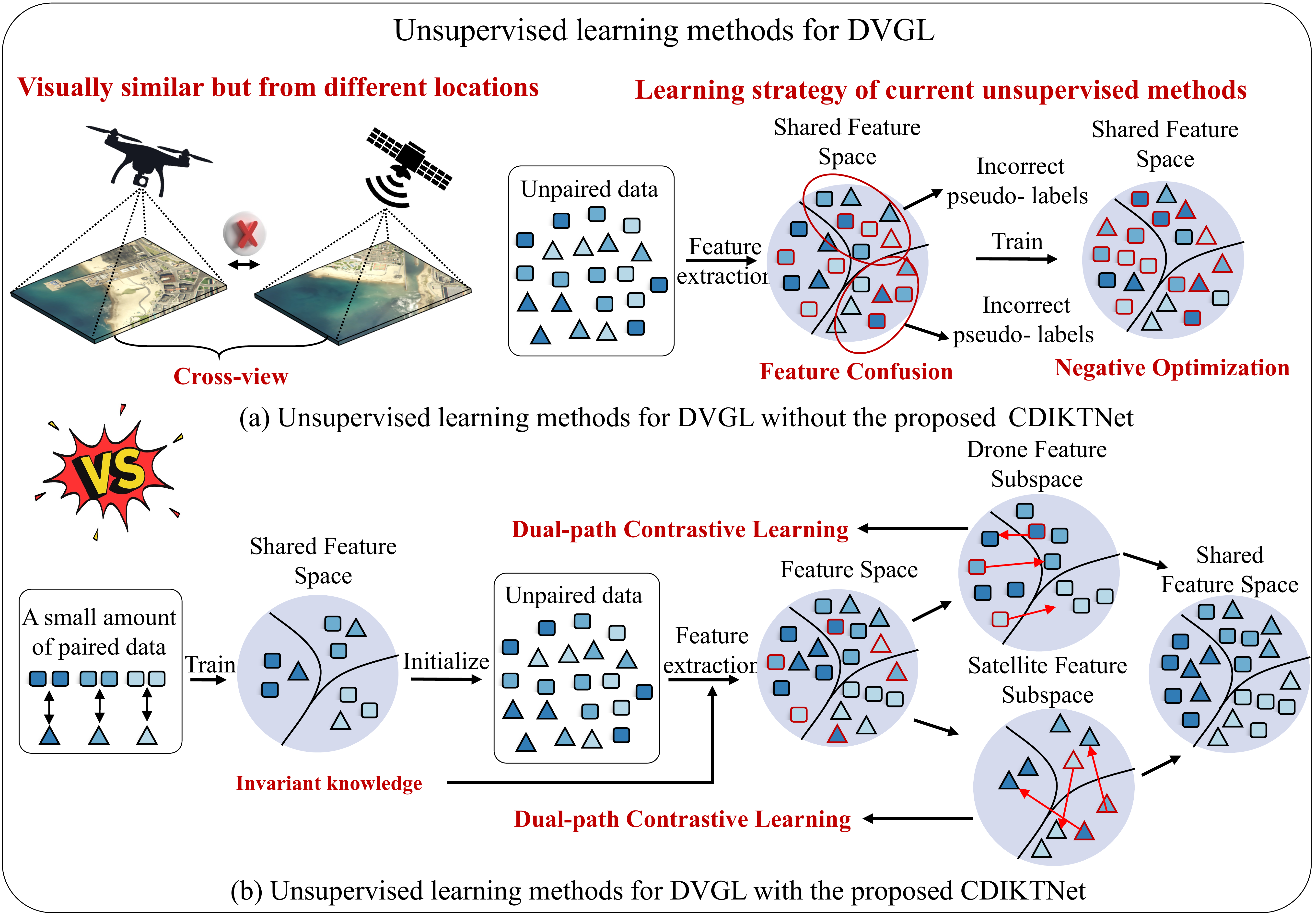}
  \caption{{\textbf{CDIKTNet vs. Existing Unsupervised Methods}}. (a) Spatial continuity causes visually similar patterns from different locations. Unsupervised methods often encounter feature confusion, which results in inaccurate pseudo-labels and drives negative optimization. (b) CDIKTNet leverages a small amount of paired data at initialization to learn cross-view invariance as prior knowledge. Based on this, it obtains a reliable shared feature space and further employs dual-path contrastive learning to optimize each subspace.}
  \label{fig2}
\end{figure}

However, as illustrated in Fig. \ref{fig1} (A(a)) and (B(a)), existing supervised methods \cite{shen2023mccg,du2024ccr,xia2024enhancing} strictly rely on a large amount of paired data to learn cross-view correlations. The lack of sufficient paired data leads to a dramatic decline in model performance. When the supervised ratio drops to 0, or in other words, when there is no paired data and all the data are unpaired, the existing supervised methods become ineffective. We mark this situation with a red X in the figures. Furthermore, dynamic drone mission environments often lead to domain shifts, which in turn require repeated efforts to re-establish cross-view correlations before deployment in new domains. It hinders the applicability in scenarios where pairing relationships are difficult to obtain and cross-domain adaptation is required. As shown in Fig. \ref{fig1} (B(a)), the new domain where all data are unpaired cannot directly employ the trained model from the source domain. Therefore, researchers have increasingly focused on unsupervised methods for DVGL. Recent work \cite{11010141} constructed a sparse 3D point cloud from multi-view aerial images and rendered pseudo-satellite images with near-orthographic views via 3D Gaussian splatting to realize geometric transformations of drone-view images. These synthetic images are then matched against real satellite-view images to enable unsupervised training. However, this method is sensitive to 3D reconstruction and pose estimation errors, and it is difficult to bridge the inherent cross-view domain gaps. In contrast, another work \cite{10589921} utilizes a frozen pre-trained backbone and a learnable adapter to align features from different viewpoints in a shared feature space. The shared feature space serves as a unified latent representation in which drone and satellite views can be directly aligned.  Cross-view pseudo-labels are generated automatically based on feature similarity as supervised signals to guide training. This method avoids explicit geometric transformations and aims to intrinsically learn cross-view correlations. Nevertheless, as shown in Fig. \ref{fig2} (a), when visually similar patterns appear at different geographical locations, the resulting feature confusion in the shared feature space often leads to incorrect pseudo-label generation. This ultimately drives the model toward negative optimization.

In this paper, we propose an efficient learning method with limited supervision for DVGL that mitigates the adverse effects of inaccurate pseudo-labels caused by feature confusion, called the cross-domain invariant knowledge transfer network (CDIKTNet). Specifically, the CDIKTNet consists of two core components, including the cross-domain invariance sub-network (CDIS) and the cross-domain transfer sub-network (CDTS). The CDIS utilizes a small amount of paired data to learn latent cross-view structural and spatial invariance, which serves as prior knowledge. As illustrated in Fig. \ref{fig2}(b), it endows the shared feature space of unpaired data with similar implicit cross-view correlations at initialization to alleviate feature confusion. Based on this, the CDTS introduces dual-path contrastive learning to independently optimize subspaces under different views. Despite being optimized separately, these subspaces maintain alignment in the shared feature space due to the shared invariance prior. The main contributions can be summarized as follows.

\begin{itemize}
  \item We propose a cross-domain invariant knowledge transfer network, which can leverage few-shot guided cross-view invariance learning to further explore latent cross-view correlations within unpaired data. It reduces reliance on strictly paired data and enables direct transfer to new domains without requiring any prior pairing information.
  \item The CDIS is designed to learn the structural and spatial invariance by integrating channel-wise local and global structural information, and performing joint interaction learning along the different dimensions of the features. It is utilized to construct a shared feature space with better compactness for subsequent unsupervised learning.
  \item The CDTS is built on the shared cross-view invariance prior, which enables unpaired data to form initial cross-view correlations in the shared feature space. On this basis, dual-path contrastive learning is employed to perform clustering and assign pseudo-labels within subspaces, which are then used as supervised signals to optimize the representations of each view.
  \item Three experimental settings are designed to encompass a spectrum of learning paradigms that span fully supervised learning and limited-supervision learning. In the context of limited supervised learning, the proportion of paired data and unpaired data is allowed for random variation. Specifically, the cross-domain initialization setting provides an effective learning paradigm for new domains without requiring any prior pairing information. Extensive experiments demonstrate that CDIKTNet outperforms existing supervised methods under full supervision and surpasses existing unsupervised methods in both few-shot initialization and cross-domain initialization settings.
    
\end{itemize}

Notably, as illustrated in Fig. \ref{fig1}(A(b)) and (B(b)), compared with supervised methods, CDIKTNet remains effective at varying levels of supervision and can be transferred to new domains without requiring paired data. Furthermore, as shown in Fig. \ref{fig2}, compared with unsupervised methods, CDIKTNet performs initial feature space optimization using limited paired data. Meanwhile, it adopts dual-path contrastive learning to optimize each subspace, which reduces feature confusion. 

The remainder of this paper is organized as follows. Section \ref{sec:Related Work} systematically reviews previous research. In Section \ref{method}, the proposed method is presented in detail. The experimental results are reported and analyzed in Section \ref{result}. Finally, conclusions are outlined in Section \ref{conclusion}.

\section{Related Work}
\label{sec:Related Work}

\subsection{Supervised Cross-view Geo-localization}\label{Cross-view Geo-localization}

Cross-view geo-localization (CVGL) is widely recognized as an image retrieval problem based on heterogeneous visual data. However, challenges arise from geometric-semantic heterogeneity and feature distribution shifts across different viewpoints. In traditional CVGL sub-tasks, such as ground-view geo-localization \cite{Liu_2019_CVPR,Hu,Shi_Yu_Liu_Zhang_Li_2020}, many researchers have employed explicit geometric alignment methods \cite{shi2019spatial,zhang2023cross,10601183} to enforce geometric consistency at varying viewpoints, while others \cite{regmi2019bridging,Toker_2021_CVPR} have used GANs \cite{goodfellow2014generative} to map source domain images to the target domain, which aims to reduce appearance discrepancies and maintain style consistency. Moreover, in drone-view geo-localization, which is another sub-task in CVGL, PCL \cite{tian2021uav} attempted to align drone-view and satellite-view images using perspective projection transformation. It also employed GAN to synthesize satellite-style images in order to reduce cross-view discrepancies and enhance both geometric and style consistency. But it was difficult to handle non-rigid deformations caused by dynamic UAV viewpoints and remained constrained by the synthetic-to-real domain gap. Moreover, such explicit feature consistency mappings fail to capture intrinsic cross-view invariance that differs from these environmentally sensitive features \cite{chen2024multi,cui2021cross}. Recently, researchers \cite{ge2024multibranch,lv2024direction} have begun to model latent structures or spatial invariance to enhance the robustness of DVGL. Nevertheless, these methods often focused solely on optimizing structures or spatial representations while neglecting their synergistic relationships. This results in limited generalization under scale variations and spatial shifts in the DVGL. Furthermore, these methods still rely on large amounts of paired data for training, which restricts their applicability in open environments.

Inspired by the aforementioned supervised learning methods, our method utilizes a small amount of paired data for initialization, and further exploits structural and spatial invariance under limited supervision to learn robust and generalizable cross-view representations. Compared with these supervised methods, our method significantly reduces the reliance on paired data while achieving superior performance.

\subsection{Unsupervised Cross-view Retrieval}

{\color{black} Supervised methods for cross-view retrieval have achieved remarkable progress \cite{9963608,10897882}. Early cross-view retrieval tasks, exemplified by person re-identification \cite{10812860,10855601,10314802,ye2021deep}, along with the current ground-view geo-localization \cite{Hu,Shi_Yu_Liu_Zhang_Li_2020} and drone-view geo-localization tasks \cite{chen2024multi,ge2024multibranch} that have been attracting extensive attention, effectively establish the instance-level feature consistency through aligning directly based on feature similarity in supervised scenarios.} However, the reliance on large-scale annotated data has limited its applicability. {\color{black} Fortunately, recent advances in domain adaptation \cite{10962298,liu2025nonconvex,li2022cross} and semi-unsupervised learning \cite{sun2021semi,jiang2022semi,du2021cross} have demonstrated that robust representations can be learned even under weak or missing labels. This research trend provides important insights for extending cross-view retrieval toward unsupervised settings and has motivated researchers to explore unsupervised cross-view retrieval methods \cite{10109197,9783116,9891821}.} Current methods \cite{Chen_2021_ICCV,Xuan_2021_CVPR,dai2022cluster} utilized powerful feature backbone networks for feature initialization, employing clustering algorithms \cite{ester1996density} to generate pseudo-labels and feature queues that provide stable negative samples for optimizing feature learning. However, these methods are primarily designed for homogeneous data with strong correlations, and their effectiveness tends to degrade significantly when applied to heterogeneous scenarios. To address the limitations, some researchers \cite{yang2022augmented,yang2024shallow} have proposed cross-modal person retrieval methods based on dual-path contrastive learning. Nevertheless, these methods still depend on pre-trained person retrieval models for feature extraction during initialization.
   
To better learn cross-view correlations, we adopt a pseudo-label-based training scheme using unpaired data, where pseudo-labels are generated through clustering in accordance with the methods described above. In contrast, the model is initialized with a small amount of paired data. It leads to a more compact and discriminative feature space and enables more reliable pseudo-label generation. Notably, although unsupervised DVGL can also be considered an unsupervised cross-domain retrieval task, existing methods remain limited and have already been discussed in the Introduction. We refrain from repeating the details here.

\section{Proposed Method}\label{method}
\textbf{Problem Formulation}. Given a drone-view image set \( \mathcal{X}^d = \{x_i^d\}_{i=1}^N \) and a satellite-view image set \( \mathcal{X}^s = \{x_j^s\}_{j=1}^M \), where each drone-view image \( x_i^d \) is paired with a corresponding satellite-view image \( x_j^s \) as a positive pair \( (x_i^d, x_j^s) \), the objective is to learn a shared feature space \( \mathcal{V} \) that minimizes the distance between positive pairs while maximizing the distance between negative pairs. Existing methods heavily rely on a large amount of paired data as positive samples for training to achieve high performance. In contrast, our method fundamentally differs by mitigating this dependency. The following sections provide a detailed explanation of our method.

\begin{figure*}[t]
  \centering
  \includegraphics[width=7.0in]{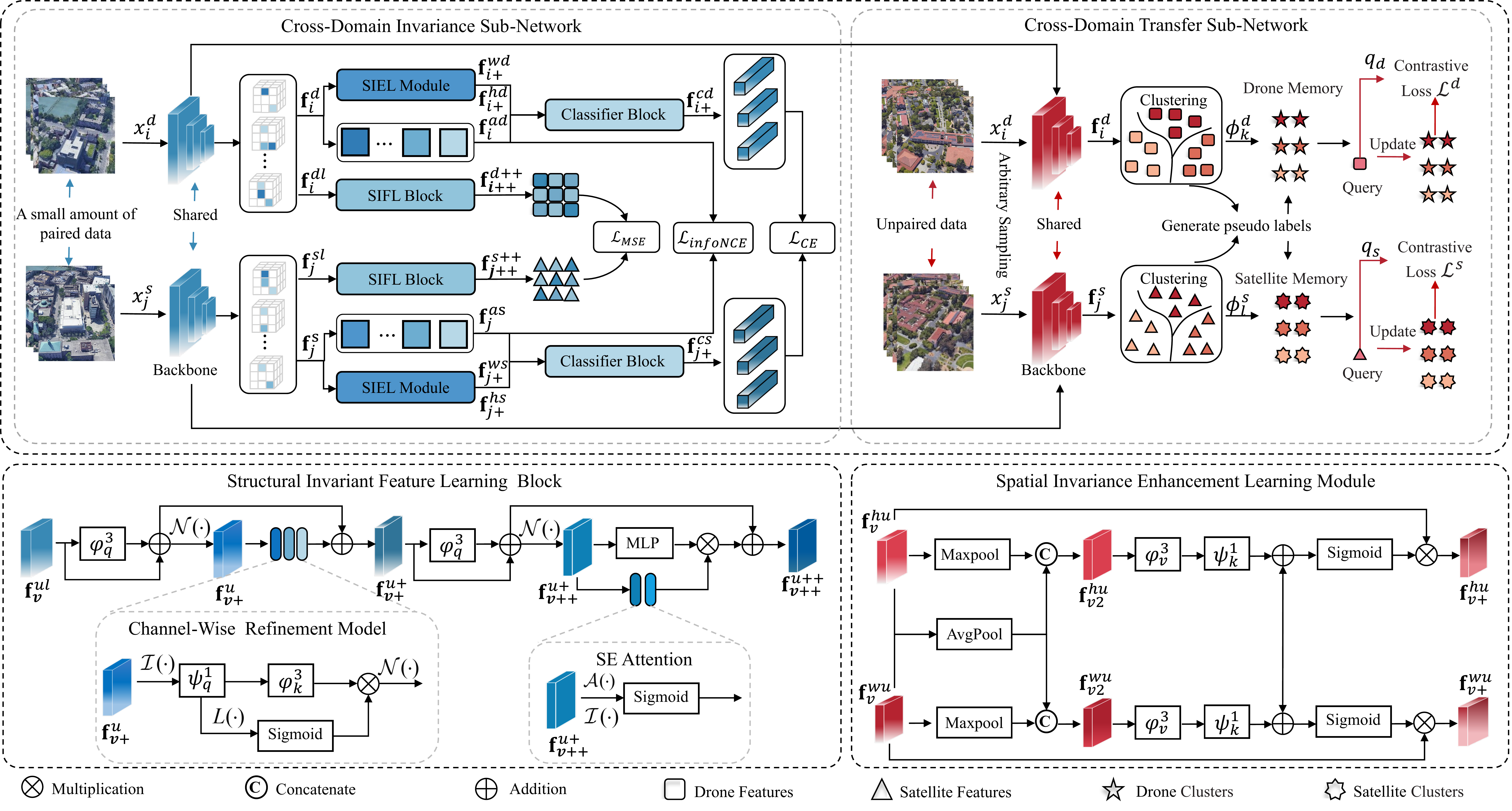}
  \caption{\textbf{Pipeline Overview}. The proposed network pipeline consists of two sub-networks. The cross-domain invariance sub-network is designed to learn feature representations that exhibit  structural invariance and spatial invariance. The cross-domain transfer sub-network utilizes the invariant knowledge learned by the cross-domain invariance sub-network as a feature representation space to further explore the latent cross-view correlations from unpaired data.}
  \label{fig3}
  \end{figure*}
\subsection{Cross-Domain Invariance Sub-network}
The CDIS lies in the complementarity between structural and spatial invariance. In DVGL, variations in viewpoint and scale are significant. Therefore, it is necessary to learn the structural and spatial features that are invariant to varying viewpoints and scales. This structural invariance consists of local geometric information, like corners, edges, and contours, and global structural information, such as spatial layout and domain-invariant topologies. Additionally, spatial invariance is derived from the scale-aware contextual cues.

\subsubsection{Structural Invariant Learning (SIFL) Block}
Given a pair of cross-view images \( \{ x_i^d, x_j^s\} \), high-dimensional feature representations are first extracted using the ConvNeXt-Base \cite{liu2022convnet} backbone network. The extracted features are represented as \( \mathbf{f}_v^u \in \mathbb{R}^{C \times H \times W} \), where \( u \in \{d, s\} \).   Specifically, \( v \)  is defined as follows: when \( u\) is \( d\) , \(v\) is \(i\) and when \(u\) is \(s\), \( v\) is \(j \). To effectively leverage structural features at different viewpoints, the dimensions \( H \times W \) are flattened into a single dimension \( L\) to obtain the features \( \mathbf{f}_{v}^{ul} \in \mathbb{R}^{C \times L} \). \( \mathbf{f}_{v}^{ul}\) serves as the input to the SIFL block. 

Subsequently, a depthwise separable \( 3 \times 3 \) convolution \( \varphi^3_q \) is used to mine local structural priors, while residual connections preserve the original topological information. After normalization \( \mathcal{N}(\cdot)\), the feature representation \(\mathbf{f}_{v+}^{u} \) can be obtained. To further enhance the capability of structural invariance feature learning, we proposed a channel-wise refinement (CWR) module. Specifically, we project \( \mathbf{f}_{v+}^{u} \) into the latent space through two cascaded linear transformation layers \( \mathcal{I}(\cdot) \)  to learn high-order interactions between channels. A subsequent \( 1 \times 1 \) convolution \( \psi^1_q \) is applied to explicitly learn local structural features. 

Following this, a gating mechanism \( \mathcal{G}(\cdot)\) consists of a depthwise separable \( 3 \times 3 \) convolution $\varphi_k^3$ branch  and a linear transformation branch $L(\cdot)$, which is performed the channel-wise selective enhancement via a sigmoid activation. Then, the residual connection and normalization \( \mathcal{N}(\cdot)\) are employed to fuse cross-layer information to enhance representation. This process can be formulated as
{\color{black}
\begin{equation}
\mathbf{f}_{v+}^{u} = \mathcal{N}\left(\varphi^3_q (\mathbf{f}_v^{ul}) + \mathbf{f}_v^{ul}\right)
\label{eq1}
\end{equation}
\begin{equation}
\mathbf{f}_{v+}^{u+} = \mathcal{N}\left(\mathcal{G}\left(\psi^1_q\left(\mathcal{I}\left(\mathbf{f}_{v+}^{u}\right)\right)\right)\right) + \varphi^3_q \left(\mathbf{f}_v^{ul}\right) + \mathbf{f}_v^{ul}
\end{equation}}

To further enhance local geometric information, we adopt the same feature processing procedure as in Eq. \ref{eq1} to extract the feature \( \mathbf{f}_{v++}^{u+} \). Although \( \mathbf{f}_{v++}^{u+} \) contains rich local geometric information, it still lacks global structural features. Therefore, we replace the CWR module described previously with an MLP \( \mathcal{P}(\cdot)\) combined with the squeeze-and-excitation (SE) attention \cite{hu2018squeeze}. In this process, the global average pooling \( \mathcal{A}(\cdot) \) is first applied within the SE attention to extract the global information at the channel level. The channel information is then compressed through two linear transformation layers \( \mathcal{I}^{+}(\cdot) \), followed by the channel attention weights calculated using the sigmoid function \(\mathcal{F}\), which are then combined with the global channel information obtained through the MLP. This further enhances the expression of global channel-level information. Finally, a residual connection is used to fuse the rich local channel-level information from the original features with the enhanced global information. This process can be formulated as
\begin{equation}
\mathbf{f}_{v++}^{u+} = \mathcal{N}\left(\varphi^3_q \left(\mathbf{f}_{v+}^{u+}\right) + \mathbf{f}_{v+}^{u+}\right)
\end{equation}
\begin{equation}
\mathbf{f}_{v++}^{u++} \!\!=\!\! \mathcal{F}\left(\mathcal{I}^{+}\left(\mathcal{A}\left(\mathbf{f}_{v++}^{u+}\right)\right)\right) \times \mathcal{P}\left(\mathbf{f}_{v++}^{u+}\right)+\varphi^3_q \left(\mathbf{f}_{v+}^{u+}\right) + \mathbf{f}_{v+}^{u+}
\end{equation}

\subsubsection{Spatial Invariant Enhancement Learning (SIEL) Module}
Spatial information is particularly crucial for DVGL. To enhance spatial invariance, inspired by \cite{shen2023mccg}, we design a dual-path spatial invariance enhancement learning module to mine spatial information from different dimensions.

{\color{black} In the spatial invariance enhancement learning module, we preserve the dimensions, namely height $H$ and width $W$ of the input features.} By permuting the input features $\mathbf{f}_v^u \in \mathbb{R}^{C \times H \times W}$ along the channel-height (C-H) and channel-width (C-W) dimensions, we obtain $\mathbf{f}_v^{hu} \in \mathbb{R}^{H \times C \times W}$ and $\mathbf{f}_v^{wu} \in \mathbb{R}^{W \times H \times C}$, respectively. This allows the module to capture multi-perspective dependencies from different dimensions. Subsequently, global feature representations $\mathbf{f}_{v2}^{hu} \in \mathbb{R}^{ 2 \times C \times H}$ and $\mathbf{f}_{v2}^{wu} \in \mathbb{R}^{ 2 \times W \times C}$ are generated by concatenating the results of max pooling and average pooling applied to the permuted features. A $3 \times 3$ depthwise separable convolution \( \varphi^3_v \) and a separate \( 1 \times 1 \) convolution \( \psi^1_k\) are then employed to obtain a more compact feature representation. Subsequently, cross-interaction across the C-H and C-W is performed to further enhance the spatial invariance of the feature representation. Finally, the fused features are normalized using a sigmoid function \(\mathcal{F}\) to generate attention weights, which are multiplied with the original feature maps and restored to the original dimensions $\mathbb{R}^{C \times H \times W}$, yielding the spatially enhanced feature $\mathbf{f}_{v+}^{hu} \in \mathbb{R}^{C \times H \times W}$ and $\mathbf{f}_{v+}^{wu} \in \mathbb{R}^{C \times H \times W}$ as
{\color{black}
\begin{align}
\mathbf{f}_{v+}^{hu} &=\mathcal{F}\left(\psi^1_k\left(\varphi^3_v\left(\mathbf{f}_{v2}^{hu}\right)\right)+\psi^1_k\left(\varphi^3_v\left(\mathbf{f}_{v2}^{wu}\right)\right)\right) \times\mathbf{f}_v^{hu} \\
\mathbf{f}_{v+}^{wu} 
&=\mathcal{F}\left(\psi^1_k\left(\varphi^3_v\left(\mathbf{f}_{v2}^{wu}\right)\right)+\psi^1_k\left(\varphi^3_v\left(\mathbf{f}_{v2}^{hu}\right)\right)\right) \times\mathbf{f}_v^{wu}
\end{align}}

{\color{black}\subsubsection{Multi-level Optimization}
  In the CDIS, the features \(\mathbf{f}_{v++}^{u++}\) obtained from the SIFL, which denotes as \(\mathbf{f}_{i++}^{d++}\) for drone-view intermediate features and  \(\mathbf{f}_{j++}^{s++}\) for satellite-view intermediate features, are optimized using the mean squared error loss \(\mathcal{L}_{MSE}\) \cite{xia2024enhancing}. Furthermore, to enhance structural representation in the spatial invariance enhancement learning module, the features \( \mathbf{f}_i^{ad} \) and \( \mathbf{f}_j^{as} \) are obtained by applying global average pooling to \( \mathbf{f}_i^{d} \) and \( \mathbf{f}_j^{s} \) , respectively. The feature  \( \mathbf{f}_i^{ad} \)  combined with \( \mathbf{f}_{i+}^{hd} \) and \( \mathbf{f}_{i+}^{wd} \) is passed through a classifier module \cite{shen2023mccg} to obtain \( \mathbf{f}_{i+}^{cd} \), as well as  \( \mathbf{f}_{j+}^{cs} \)  obtained in the same manner.  Subsequently, \( \mathbf{f}_{i+}^{cd} \) and \( \mathbf{f}_{j+}^{cs} \) are optimized with cross-entropy loss \( \mathcal{L}_{CE} \) to enhance feature discrimination. In addition, to further enhance the feature representations, \( \mathbf{f}_i^{ad}\) and  \( \mathbf{f}_j^{as}\) undergo independent optimization using InfoNCE loss \(\mathcal{L}_{InfoNCE}\) \cite{oord2018representation}. The final objective is to minimize the sum of all losses. It can be formulated as
\begin{equation}
   \mathcal{L}_{1} = \lambda_1\mathcal{L}_{MSE} + \lambda_2\mathcal{L}_{CE}+ \lambda_3\mathcal{L}_{infoNCE}
    \label{eq7}
\end{equation}
where \(\lambda_1\), \(\lambda_2\) and \(\lambda_3\) serve as loss coefficients to balance the relative contributions of individual loss terms.}

\remark{The CDIS lies in dual optimization of structural and spatial invariance. On one hand, it integrates local and global structural information along the channel dimension and is guided by \(\mathcal{L}_{MSE}\) to effectively constrain the feature search space, which enhances the structural representations. On the other hand, the CDIS retains the original spatial dimensions and performs joint interaction learning along C-H and C-W, where \( \mathcal{L}_{CE} \) is employed to further mine scale-aware contextual cues. This structure–spatial collaborative modeling enables CDIS to achieve rapid convergence and learn highly discriminative representations, even with limited paired data.}

\subsection{Cross-Domain Transfer Sub-network}
The CDTS further explores cross-view correlations based on the enhanced compactness established by CDIS. Specifically, the CDTS incorporates dual-path contrastive learning tailored explicitly for DVGL.
 
\subsubsection{Cross-View Feature Memory Initialization}
{\color{black} As shown in Fig. \ref{fig3}, at the beginning of each training epoch, cross-view feature instances are initialized using the invariant knowledge learned by CDIS and then clustering is performed to generate pseudo-labels. It can mitigate the issue of low-quality pseudo-label initialization caused by suboptimal feature extraction. Subsequently, cluster memories are constructed for the drone-view and satellite-view. These memories $\{\phi_k^d\}_{k=1}^{K}$ and $\{\phi_l^s\}_{l=1}^{L}$ are obtained by computing the average feature representation of all drone or satellite instances within each cluster, where $K$ and $L$ denote the total number of drone-view clusters and satellite-view clusters, respectively. $\phi_k^d$ and $\phi_l^s$ serve as the cluster centroids for the drone-view and satellite-view features, respectively. This process can be formulated as
\begin{equation}
\phi_k^d = \frac{1}{|\mathcal{H}_k^d|} \sum_{\mathbf{f}_i^d \in \mathcal{H}_k^d} \mathbf{f}_i^d, \phi_l^s = \frac{1}{|\mathcal{H}_l^s|} \sum_{\mathbf{f}_j^s \in \mathcal{H}_l^s} \mathbf{f}_j^s
\end{equation}
where \(\mathbf{f}_i^d \) and \(\mathbf{f}_j^s \) are drone and satellite instance features, respectively. \(i=1,\cdots,N\), \(j=1,\cdots,M\). \(\mathcal{H}_{k}^{d}\) and \(\mathcal{H}_{l}^{s}\) denote the $k$-th cluster in drone-view and $l$-th cluster in satellite-view, respectively. \(|\cdot|\) indicates the number of instances per cluster.}

{\color{black}\remark{In selecting the clustering algorithm, we refer to several unsupervised cross-view retrieval works [\cite{dai2022cluster,adca} and adopt DBSCAN \cite{ester1996density} for clustering. Unlike partition-based methods such as K-means \cite{likas2003global}, DBSCAN utilizes a density-based method, which handles clusters of arbitrary shapes and noisy data. It does not require a predefined number of clusters, nor does it force all points to be assigned to a cluster. Additionally, DBSCAN can identify and exclude noise points, which ensures the selection of valid samples even in cross-view challenges.}}

\subsubsection{Cross-View Feature Memory Updating}
During each training iteration, the cluster memories using a momentum updating strategy are represented as
\begin{equation}
\quad \phi_k^{d(\omega)} \leftarrow \alpha \phi_k^{d(\omega-1)} + (1 - \alpha) q_d
\label{eq10}
\end{equation}
\begin{equation}
\quad \phi_l^{s(\omega)} \leftarrow \alpha \phi_l^{s(\omega-1)} + (1 - \alpha) q_s
\label{eq11}
\end{equation}
where \(\alpha\) is the momentum factor and \(\omega\) is the iteration number.
\( q_d\) and \( q_s\) represent the given query features of the drone and the satellite, respectively.
{\color{black}
\subsubsection{Cross-View Contrastive Loss}
Given the drone and satellite query \( q_d\) and \( q_s\), we compute the contrastive loss for the drone view and the satellite view as
\begin{equation}
\mathcal{L}^{d} = - \log \frac{\exp(q_{d} \cdot \phi^{d}_{+} / \tau)}{\sum_{k=1}^{K} \exp(q_{d} \cdot \phi^{d}_k / \tau)} \tag{11}
\end{equation}
\begin{equation}
\mathcal{L}^{s} = - \log \frac{\exp(q_{s} \cdot \phi^{s}_{+} / \tau)}{\sum_{l=1}^{L} \exp(q_{s} \cdot \phi^{s}_l / \tau)} \tag{12}
\end{equation}
where  \(\tau\) denotes the temperature hyper-parameter. \(\phi^{d}_{k}\) (or \(\phi^{s}_{l}\) ) is utilized as a feature vector at the cluster level to compute the distances between the query instance \( q_d\) (or \( q_s\)) and all clusters. 
 Here, \(\phi^{d}_+\) and \(\phi^{s}_+\) correspond to the positive cluster memory associated with \(q_{d}\) and \(q_{s}\) in \(\{\phi_k^d\}_{k=1}^{K}\) and \(\{\phi_l^s\}_{l=1}^{L} \), respectively. In practice, for a given query feature, e.g., $q_{d}$ or $q_{s}$, the cluster memory that is assigned in the same cluster, namely $\phi^{d}_{+}$ or $\phi^{s}_{+}$, is designated as the positive sample.}

\remark{Benefiting from CDIS, instances \(\mathbf{f}_i^d\) and \(\mathbf{f}_j^s\) exhibit stronger compactness within their respective subspaces and cross-view correlation in shared feature space. This also makes \(\phi_k^d\) and \(\phi_l^s\) more representative and an initial alignment in the shared feature space. On this basis, \(\mathcal{L}^{d}\) and \(\mathcal{L}^{s}\) further guide each instance feature towards its corresponding \(\phi_k^d\) and \(\phi_l^s\) within the subspace, thereby indirectly facilitating cross-view alignment in the shared feature space.}

The CDTS is optimized jointly with \(\mathcal{L}^{d}\) and \(\mathcal{L}^{s}\). Thus, the final optimization for CDIKTNet is denoted as \(\mathcal{L} \!= \!\mathcal{L}_{1}\! + \!\mathcal{L}^{d} \!+\! \mathcal{L}^{s}\).

\section{Experimental Results}\label{result}
\subsection{Datasets and Evaluation Protocol}
CDIKTNet is evaluated on two challenging benchmarks for DVGL. University-1652 dataset \cite{zheng2020university} consists of drone, satellite, and ground-view images from 1,652 locations in 72 universities. Its training set includes 701 buildings from 33 universities, while the test set comprises 951 buildings from 39 universities. SUES-200 dataset \cite{zhu2023sues} comprises 200 locations, which are split into 120 for training and 80 for testing. Each location provides a satellite-view image and drone-view images taken at four different altitudes, which encompass diverse environments such as parks, lakes, and buildings. We employ Recall@K (R@K) and Average Precision (AP) as evaluation metrics. Additionally, Ground Truth Ratio (GT ratio) denotes the proportion of paired data within the entire experimental dataset.

\subsection{Implementation Details}
During training and testing, all input images are resized to \( 3 \times384 \times384 \). The CDIS is trained for 1 epoch, while the CDTS runs for 30 epochs, with a batch size of 64. In DBSCAN, the neighborhood radius is set to 0.40 and 0.30 for drone-view images and satellite-view images, respectively, and the minimum number required to form a cluster is 4.  AdamW optimizer is used for CDIS with an initial learning rate of 0.001, followed by SGD optimization for the CDTS with a learning rate of 0.00025. Furthermore, in Eq.(\ref{eq7}), \(\lambda_1\), \(\lambda_2\), and \(\lambda_3\) are set to 0.6, 0.1, and 1, respectively. The CDTS is activated only when there exist unpaired data.

{\color{black}
\subsection{Comparison with State-of-the-art Methods}
We conduct experiments with three settings, namely (\rmnum{1}) fully supervised learning, (\rmnum{2}) few-shot initialization, and (\rmnum{3}) cross-domain initialization. The methods in comparison encompass a range of state-of-the-art methods. For supervised learning, these include MuSe-Net \cite{wang2024multiple}, TransFG \cite{zhao2024transfg}, IFSs \cite{ge2024multibranch}, MCCG \cite{shen2023mccg}, Sample4Geo \cite{deuser2023sample4geo}, CAMP \cite{wu2024camp}, DAC \cite{xia2024enhancing}, QDFL \cite{hu2025query} and CDM-Net \cite{zhou2025cdm}. In the realm of unsupervised learning, the methods are  EM-CVGL \cite{10589921}, SegVLAD \cite{garg2024revisit}, AnyLoc \cite{keetha2023anyloc} and Li \cite{11010141}. Notably, SegVLAD \cite{garg2024revisit}, AnyLoc \cite{keetha2023anyloc} and Li \cite{11010141} are reported in separate Tables \ref{tab:university1652}, \ref{tab:sues-200} and \ref{tab:sues-20012} due to the absence of complete results in their publications.}

{\color{black}
\textbf{(\rmnum{1}) Fully Supervised Learning}: Compared with existing state-of-the-art methods, CDIKTNet in this setting achieves competitive performance on both datasets. Specifically, on University-1652, it attains 95.02\% R@1 and 95.79\% AP in the drone$\rightarrow$satellite evaluation scenario. Overall, it outperforms advanced supervised methods such as DAC \cite{xia2024enhancing} and CAMP \cite{wu2024camp}, while achieving performance comparable to that of the state-of-the-art methods QDFL \cite{hu2025query} and CDM-Net \cite{zhou2025cdm}. On SUES-200, across two evaluation scenarios and four altitudes, CDIKTNet generally outperforms advanced supervised methods such as DAC \cite{xia2024enhancing} and CAMP \cite{wu2024camp}, and significantly surpasses Sample4Geo \cite{deuser2023sample4geo}, while also showing comparable performance to QDFL \cite{hu2025query} and CDM-Net \cite{zhou2025cdm}. Furthermore, as shown in Table~~\ref{tab:university-seus}, CDIKTNet trained in this setting demonstrates superior generalization capability compared with advanced supervised methods when models trained on University-1652 are directly tested on SUES-200. These results validate the effectiveness of CDIKTNet for fully supervised learning.}

\begin{table}[t]
\caption{Comparisons between existing state-of-the-art methods and CDIKTNet on University-1652.}
\centering
\tiny
\resizebox{\columnwidth}{!}{ 
\setlength{\tabcolsep}{3pt}
\begin{tabular}{ccccccc}\hline
\multicolumn{7}{c}{University-1652}  \\ \hline
\multirow{2}{*}{Setting} &\multirow{2}{*}{Model} & \multirow{2}{*}{\makecell[c]{GT Ratio\\(\%)}} & \multicolumn{2}{c}{Drone$\rightarrow$Satellite} & \multicolumn{2}{c}{Satellite$\rightarrow$Drone} \\ \cline{4-7} 
&  & & R@1 & AP   & R@1 & AP\\ \hline
\multirow{10}{*}{\rmnum{1}} &  MuSe-Net\cite{wang2024multiple}    &100        & 74.48       & 77.83      & 88.02     & 75.10 \\
&TransFG\cite{zhao2024transfg}       & 100   & 84.01       & 86.31      & 90.16     & 84.61 \\
&  IFSs\cite{ge2024multibranch}       &100        & 86.06       & 88.08      & 91.44     & 85.73 \\
&  MCCG\cite{shen2023mccg}           &100 & 89.40       & 91.07      & 95.01     & 89.93 \\
&  Sample4Geo\cite{deuser2023sample4geo}  &100       & 92.65       &93.81       & 95.14     & 91.39 \\
&  CAMP\cite{wu2024camp}                &100  &{94.46}   &{95.38}    &{96.15}     & 92.72 \\
&  DAC\cite{xia2024enhancing}          &100  &{94.67} &{95.50} &{96.43}&{93.79} \\ 
&  QDFL\cite{hu2025query}          &100  &{95.00} &{95.83} &{97.15}&{94.57} \\ 
&  CDM-Net\cite{zhou2025cdm}          &100  &{95.13} &{96.04} &{96.68}&{94.05} \\ 
&CDIKTNet  &100 &\textbf{95.02}       &\textbf{95.79} &\textbf{96.15}     & \textbf{93.93}\\\hline
\multirow{2}{*}{\rmnum{2}} & CDIKTNet   &10 &{88.02}       &{89.84} &{93.87}     & {86.48} \\
&CDIKTNet     &2 &{85.63}       &{87.75} &{90.44}     & {82.24}\\\hline
UM & EM-CVGL\cite{10589921}  &0  & 70.29      & 74.93       & 79.03      & 61.03    \\\hline
\multirow{1}{*}{\rmnum{3}}& CDIKTNet    &2/0 &\textbf{83.30}       &\textbf{85.73} &\textbf{87.73}     & \textbf{76.53}\\ \hline
\end{tabular}}
\label{tab:university}
\footnote{1}{\footnotesize “UM” is employed as the abbreviated form of “Unsupervised Method”.\qquad\qquad\qquad\qquad\qquad\qquad\qquad\qquad\qquad}
\end{table}

\begin{table*}[t]
\centering
\tiny
\caption{Comparisons between existing state-of-the-art methods and CDIKTNet on SUES-200.}
  \setlength{\tabcolsep}{3pt}
\resizebox{\textwidth}{!}{%
\begin{tabular}{ccccccccccc|cccccccc}
\hline
&\multicolumn{17}{c}{SUES-200} \\ \hline
\multirow{3}{*}{Setting} & \multirow{3}{*}{Model} & \multirow{3}{*}{\makecell[c]{GT Ratio\\(\%)}} & \multicolumn{8}{c|}{Drone$\rightarrow$Satellite} & \multicolumn{8}{c}{Satellite$\rightarrow$Drone} \\ \cline{4-19} 
               &        &                                & \multicolumn{2}{c}{150m} & \multicolumn{2}{c}{200m} & \multicolumn{2}{c}{250m} & \multicolumn{2}{c|}{300m} & \multicolumn{2}{c}{150m} & \multicolumn{2}{c}{200m} & \multicolumn{2}{c}{250m} & \multicolumn{2}{c}{300m} \\ \cline{4-19} 
           &            &                                & R@1 & AP   & R@1 & AP   & R@1 & AP   & R@1 & AP   & R@1 & AP   & R@1 & AP   & R@1 & AP   & R@1 & AP \\ \hline
\multirow{8}{*}{\rmnum{1}}& IFSs \cite{ge2024multibranch}   &100 &77.57 &81.30 &89.50 &91.40 &92.58 &94.21 &97.40 &97.92 &93.75 &79.49 &{97.50} &90.52 &97.50 &96.03 &{100.00} &97.66\\ 
&MCCG\cite{shen2023mccg}     &100 &82.22 &85.47 &89.38 &91.41 &93.82 &95.04 &95.07 &96.20 &93.75 &89.72 &93.75 &92.21 &96.25 &96.14 &{98.75} &96.64\\ 
&Sample4Geo\cite{deuser2023sample4geo} &100 &92.60 &94.00 &97.38 &97.81 &{98.28} &{98.64} &99.18 &99.36 &{97.50} &93.63 &{98.75} &{96.70} &{98.75} &{98.28} &{98.75} &98.05\\
&DAC\cite{xia2024enhancing}&100 &{96.80}  &{97.54} & 97.48 & 97.97 & 98.20 & 98.62 & 97.58 & 98.14  &{97.50} &{94.06}  &{98.75} & 96.66 &{98.75} & 98.09 &{98.75} & 97.87 \\
&CAMP\cite{wu2024camp} &100&{95.40}  &{96.38} &{97.63}  &{98.16} & 98.05 & 98.45 &{99.33} &{99.46} &{96.25} &{93.69} &{97.50} &{96.76} &{98.75} & 98.10 &{100.00} &{98.85}  \\
&CDM-Net\cite{zhou2025cdm} &100&{93.78}  &{95.16} &{97.62}  &{98.16} & 98.28 & 98.69 &{99.20} &{99.31} &{95.25} &{92.24} &{98.50} &{96.40} &{99.00} & 97.60 &{99.00} &{98.01}  \\
&QDFL\cite{hu2025query} &100&{93.97}  &{95.42} &{98.25}  &{98.67} & 99.30 & 99.48 &{99.31} &{99.48} &{98.75} &{95.10} &{98.75} &{97.92} &{100.00} & 99.07 &{100.00} &{99.07}  \\
&CDIKTNet &{100} &\textbf{96.03} &\textbf{96.87} &\textbf{97.80} &\textbf{98.27} &\textbf{98.90} &\textbf{99.15} &\textbf{99.73}   &\textbf{99.79}      &\textbf{97.50}   &\textbf{94.23}   &\textbf{98.75} &\textbf{96.58}   &\textbf{100.00}   &\textbf{98.33}   &\textbf{98.75}    &\textbf{98.84} \\ \hline
\multirow{2}{*}{\rmnum{2}} & CDIKTNet &{10} &{83.60} &{86.24} & {90.53} &{92.14} & {94.33} & {95.32} &{97.13}   &{97.60}    & {91.25}   &{83.24}   &{95.00} &{89.53}   &{97.50}   &{93.59}   &{97.50}    &{95.80} \\ 

&CDIKTNet &{2} &{80.55} & {83.58} &{88.05} &{90.10} & {91.28} &{93.13} &{94.48}  &{96.12}    &{90.00}   &{77.16}   &{91.25} &{85.92}   &{97.50}   &{91.36}   &{97.50}    &{94.07}   \\\hline

\multirow{1}{*}{UM}& EM-CVGL\cite{10589921}&0 & 55.23      & 60.80       & 60.95      & 61.03 & 68.10      & 72.62       & 74.42      & 78.20 & 73.75      &54.00       & 91.25      & 65.65  & 96.25      & 72.02       & 97.50      & 74.74\\  \hline
\multirow{1}{*}{\rmnum{3}}& CDIKTNet &{2/0} &\textbf{82.75} &\textbf {85.25} &\textbf{89.35} &\textbf{91.13} & \textbf{93.15} &\textbf{94.39} &\textbf{95.18}  &\textbf{96.12}    &\textbf{88.75}   &\textbf{80.33}   &\textbf{93.75} &\textbf{88.17}   &\textbf{95.00}   &\textbf{92.15}   &\textbf{98.75}    &\textbf{94.37}\\\hline

\end{tabular}%
}
\label{tab:SUES-200}
\end{table*}
\begin{table*}[t]
\centering
\caption{Comparisons between existing state-of-the-art methods and CDIKTNet trained in setting  \rmnum{1}\  and  \rmnum{2}\  in generalization capacity.}
\tiny
  \setlength{\tabcolsep}{3pt}
\resizebox{\textwidth}{!}{%
\begin{tabular}{ccccccccccc|cccccccc}
\hline
&\multicolumn{17}{c}{University-1652$\rightarrow$SUES-200} \\ \hline
\multirow{3}{*}{Setting} & \multirow{3}{*}{Model} & \multirow{3}{*}{\makecell[c]{GT Ratio\\(\%)}} & \multicolumn{8}{c|}{Drone$\rightarrow$Satellite} & \multicolumn{8}{c}{Satellite$\rightarrow$Drone} \\ \cline{4-19} 
                   &    &                                & \multicolumn{2}{c}{150m} & \multicolumn{2}{c}{200m} & \multicolumn{2}{c}{250m} & \multicolumn{2}{c|}{300m} & \multicolumn{2}{c}{150m} & \multicolumn{2}{c}{200m} & \multicolumn{2}{c}{250m} & \multicolumn{2}{c}{300m} \\ \cline{4-19} 
                    &   &                                & R@1 & AP   & R@1 & AP   & R@1 & AP   & R@1 & AP   & R@1 & AP   & R@1 & AP   & R@1 & AP   & R@1 & AP \\ \hline

\multirow{5}{*}{\rmnum{1}} & MCCG\cite{shen2023mccg}    &100 & 57.62 & 62.80 & 66.83 & 71.60 & 74.25 & 78.35 & 82.55 & 85.27 & 61.25 & 53.51 & 82.50 & 67.06 & 81.25 & 74.99 & 87.50 & 80.20\\

&Sample4Geo\cite{deuser2023sample4geo} &100 & 70.05 & 74.93 & 80.68 & 83.90 & 87.35 & 89.72 & 90.03 & 91.91 & 83.75 & 73.83 & 91.25 & 83.42 &{93.75} & 89.07 & 93.75 & 90.66\\

&DAC\cite{xia2024enhancing} &100 & 76.65 & 80.56 & 86.45 & 89.00 &{92.95}  & {94.18} &{94.53} & {95.45}   & {87.50} &{79.87} &{96.25} &{88.98}  &{95.00} &{92.81} &{96.25} &{94.00}   \\
&CAMP\cite{wu2024camp} &100& {78.90} &{82.38} &{86.83}  &{89.28} & 91.95 &{93.63} &{95.68}  &{96.65} &{87.50} & 78.98 &{95.00} & 87.05  &{95.00} & 91.05 &{96.25} & 93.44\\

&CDIKTNet &{100} &\textbf{80.20} &\textbf{81.20} &\textbf{88.85} & \textbf{90.65} & \textbf{93.32} &\textbf{94.51} &\textbf{94.10}   &\textbf{95.16}    &\textbf{90.00}   &\textbf{82.85}   &\textbf{93.75} &\textbf{89.92}   &\textbf{96.25}   &\textbf{92.48}   &\textbf{95.00}    &\textbf{93.62} \\ \hline
\multirow{2}{*}{\rmnum{2}} & CDIKTNet &{10} &{81.35} &{84.28} &{89.15} &{91.16} &{94.18} &{95.29} &{96.68}   &{97.32} 
&{91.25}   &{81.10}   &{97.50} &{90.44}   &{98.75}   &{93.87}   &{98.75}    &{95.05} \\ 
  & CDIKTNet  &{2} &{70.98} &{75.15} &{80.18} &{83.47} &{87.05} &{89.36} &{87.05}  &{94.37}    &{81.25}   &{71.05}   &{92.50} &{81.20}   &{92.50}   &{87.60}   &{96.25}    &{91.49}   \\ \hline
 \multirow{1}{*}{UM} & EM-CVGL\cite{10589921}  &0 & 60.03  &65.69       & 71.50      & 76.17  & 80.35  & 83.85       & 85.93      & 88.34    & 73.75  &56.99       & 87.50      & 70.62  & 92.50  & 81.18       & 95.00      & 86.04\\ \hline
 \multirow{1}{*}{\rmnum{3}} & CDIKTNet  &2/0 & \textbf{79.08}  &\textbf{82.55}  & \textbf{90.07}   & \textbf{91.85}  & \textbf{93.05}  &\textbf {94.37}   &\textbf {95.00}      & \textbf{95.95}    & \textbf{86.25}  &\textbf{77.90}       &\textbf {92.50}      & \textbf{86.40}  & \textbf{93.75}  &\textbf {90.16}       &\textbf{95.00}      & \textbf{92.22}\\ \hline
\end{tabular}%
}

\label{tab:university-seus}
\end{table*}

\begin{table}[htbp]
  \centering
  \tiny
  \caption{Comparisons between existing state-of-the-art unsupervised methods and CDIKTNet on university-1652.}
  \label{tab:university1652}
    \resizebox{\columnwidth}{!}{ 
  \begin{tabular}{ccccccc}
    \hline
    \multicolumn{7}{c}{University-1652} \\ \hline
   \multirow{2}{*}{Setting} &\multirow{2}{*}{Model}& \multirow{2}{*}{\makecell[c]{GT Ratio\\(\%)}} & \multicolumn{4}{c}{Drone→Satellite}\\ \cline{4-7}
                 & &          & R@1    & R@5    & R@10  & AP \\ \hline
    \multirow{4}{*}{UM} & AnyLoc\cite{keetha2023anyloc}       & 0        & 42.08  & 57.92  & 69.19 &- \\
    &SegVLAD\cite{garg2024revisit}      & 0        & 68.19  & 77.32  & 77.18 &- \\
&Li(a)\cite{11010141}          & 0        & 57.49  & 73.32  & 77.18 &- \\
&Li(b)\cite{11010141}           & 0        & 76.60  & 80.46  & 84.88 &- \\ \hline
       \multirow{2}{*}{\rmnum{2}} & CDIKTNet   &10 &{88.02}       &{89.84} &{93.87}     & {86.48} \\
       &CDIKTNet     & 2        & 85.63  & 94.86  & 96.31 &87.75 \\\hline
    \multirow{1}{*}{\rmnum{3}} &CDIKTNet     & 2/0        & \textbf{83.30}  & \textbf{94.00}  & \textbf{95.71} &\textbf{85.73} \\ \hline
  \end{tabular}}
  \footnote{2}{\footnotesize Li(a)\cite{11010141} and Li(b)\cite{11010141} correspond to different parameter configurations reported in the original paper.\qquad\qquad\qquad\qquad\qquad\qquad\qquad\qquad\qquad}
\end{table}

\begin{table}[]
  \centering
  \tiny
  \caption{Comparisons between existing state-of-the-art unsupervised methods and CDIKTNet on SUES-200}
  \label{tab:sues-200}
  \resizebox{\columnwidth}{!}{ 
  \begin{tabular}{ccccccc}
    \hline
    \multicolumn{7}{c}{SUES-200} \\ \hline
      \multirow{3}{*}{Setting} &\multirow{3}{*}{Model}& \multirow{3}{*}{\makecell[c]{GT Ratio\\(\%)}} & \multicolumn{4}{c}{Drone→Satellite}\\ \cline{4-7}
    & & & \multicolumn{2}{c}{200m} & \multicolumn{2}{c}{300m}\\\cline{4-7}
               & &           & R@1    & AP    & R@1  & AP \\ \hline
    \multirow{4}{*}{UM}&AnyLoc\cite{keetha2023anyloc}       & 0        & 34.00  & 46.74  & 41.00 & 51.94 \\
    &SegVLAD\cite{garg2024revisit}      & 0        & 69.50  & 72.98  & 73.00 & 74.26\\
&Li(a)\cite{11010141}           & 0        & 56.00  & 60.50  & 62.00 & 65.00 \\
&Li(b)\cite{11010141}           & 0        & 73.00  & 76.54  & 76.50 & 77.52\\\hline
\multirow{2}{*}{\rmnum{2}} &CDIKTNet     & 10     &89.15 &91.16    &96.68 &97.32 \\

&CDIKTNet     & 2        & 88.05  &90.10   & 94.48 &96.12  \\\hline
\multirow{1}{*}{\rmnum{3}} &CDIKTNet     & 2/0        & \textbf{89.35}  &\textbf{91.13}   & \textbf{95.18} &\textbf{96.12}  \\     \hline
  \end{tabular}}
\end{table}

\begin{table}[]
  \centering
  \caption{Comparisons between existing state-of-the-art unsupervised methods and CDIKTNet trained in setting \rmnum{2}~in generalization capacity.}
  \label{tab:sues-20012}
      \setlength{\tabcolsep}{3pt}
\resizebox{\columnwidth}{!}{ 
  \begin{tabular}{ccccccccccc}
    \hline
    \multicolumn{11}{c}{University-1652$\rightarrow$SUES-200} \\ \hline
   \multirow{3}{*}{Setting} &\multirow{3}{*}{Model}& \multirow{3}{*}{\makecell[c]{GT Ratio\\(\%)}} & \multicolumn{8}{c}{Drone→Satellite}\\ \cline{4-11}
    && & \multicolumn{2}{c}{150m} & \multicolumn{2}{c}{200m} & \multicolumn{2}{c}{250m} & \multicolumn{2}{c}{300m} \\\cline{4-11}
     &&          & R@1    & AP    & R@1  & AP  & R@1    & AP    & R@1  & AP\\ \hline
    \multirow{3}{*}{UM} &AnyLoc\cite{keetha2023anyloc}       & 0        &30.00	&42.74	&34.00	&46.74 &36.78	&50.41 &41.00 &51.97\\
    &SegVLAD\cite{garg2024revisit}      & 0        &56.00	&59.08	&69.50	&72.98 &72.50	&74.10 &73.00 &74.26\\
&Li(b)\cite{11010141}           & 0        &66.00	&69.98	&73.00	&76.54 &73.50	&75.15 &76.50 &77.52 \\ \hline
    
\multirow{2}{*}{\rmnum{2}} & CDIKTNet &{10} &{81.35} &{84.28} &{89.15} &{91.16} &{94.18} &{95.29} &{96.68}   &{97.32}\\
&CDIKTNet     & 2        &{70.98} & {75.15} &{80.18} &{83.47} &{87.05} &{89.36} &{87.05}  &{94.37}  \\\hline
\multirow{1}{*}{\rmnum{3}} & CDIKTNet  &2/0 & \textbf{79.08}  &\textbf{82.55}  & \textbf{90.07}   & \textbf{91.85}  & \textbf{93.05}  &\textbf {94.37}   &\textbf {95.00}      & \textbf{95.95}  \\\hline
  \end{tabular}}
\end{table}

\begin{table*}[t]
\centering
\caption{Comparisons between   existing state-of-the-art methods and CDIKTNet on SUES-200 with 2\% and 10\% paired data.}
\tiny
  \setlength{\tabcolsep}{3pt}
\resizebox{\textwidth}{!}{%
\begin{tabular}{ccccccccccc|cccccccc}
\hline
&\multicolumn{17}{c}{SUES-200} \\ \hline
\multirow{3}{*}{Setting} &\multirow{3}{*}{Model} & \multirow{3}{*}{\makecell[c]{GT Ratio\\(\%)}} & \multicolumn{8}{c|}{Drone$\rightarrow$Satellite} & \multicolumn{8}{c}{Satellite$\rightarrow$Drone} \\ \cline{4-19} 
                      &  &                                & \multicolumn{2}{c}{150m} & \multicolumn{2}{c}{200m} & \multicolumn{2}{c}{250m} & \multicolumn{2}{c|}{300m} & \multicolumn{2}{c}{150m} & \multicolumn{2}{c}{200m} & \multicolumn{2}{c}{250m} & \multicolumn{2}{c}{300m} \\ \cline{4-19} 
                      & &                                & R@1 & AP   & R@1 & AP   & R@1 & AP   & R@1 & AP   & R@1 & AP   & R@1 & AP   & R@1 & AP   & R@1 & AP \\ \hline

\multirow{3}{*}{-} &Sample4Geo\cite{deuser2023sample4geo}     &2 &{62.10} &{68.11} &{69.85} & {74.93} &{74.85} &{79.14} &{76.40} &{80.33} &{77.50} &{57.91} & {88.75} &{70.17} &{91.25} & {76.99} & {90.00} &{78.89}\\ 

&CAMP\cite{wu2024camp}&2 & 54.73 & 61.70 & 62.30 & 68.39 & 65.83 & 71.42 & 68.40 & 73.58 & 68.75 & 49.25 & 82.50 & 60.49 &{87.50} & 68.79 &{91.25} & 74.43\\

&DAC\cite{xia2024enhancing} &2 &{55.83} & {62.16} & {68.50} &{73.25} &{73.38}  &{77.34} &{74.70} &{78.64}   &{70.00} &{51.09}  &{85.00} &{61.59}  &{90.00} &{70.53} &{91.25} &{74.63}   \\\hline
\multirow{1}{*}{\rmnum{2}}&CDIKTNet &{2} &\textbf{80.55} &\textbf{83.58} &\textbf{88.05} &\textbf{90.10} &\textbf{91.28} &\textbf{93.13} &\textbf{94.48}  &\textbf{96.12}    &\textbf{90.00}   &\textbf{77.16}   &\textbf{91.25} &\textbf{85.92}   &\textbf{97.50}   &\textbf{91.36}   &\textbf{97.50}    &\textbf{94.07}   \\  \hline 

\multirow{3}{*}{-} &Sample4Geo\cite{deuser2023sample4geo} &{10} & 75.43 & 79.21 & 84.20 & 86.70 & 87.18 & 89.40 & 89.63 & 91.61 & 91.25 & 76.89 & 92.50 & 82.76 &{96.25} & 86.60 & {96.25} & 89.23\\
&CAMP\cite{wu2024camp} &{10} &{76.27} &{80.39} &{83.83} &{86.89} &{89.66} &{91.62} &{90.85} &{92.71} 
&{92.50}   &{77.50}   &{96.25} &{86.63}   &{97.50}   &{91.45}   &{97.50}    &{93.29} \\ 
&DAC\cite{xia2024enhancing}  &{10} &{76.80} & {80.65} &{85.05} &{87.65} & {88.80} &{90.81} &{90.58}  &{92.32}    &\textbf{93.75}   &{79.06}   &\textbf{97.50} &{86.78}   &\textbf{97.50}   &{90.58}   &\textbf{98.75}    &{92.54}   \\ \hline
\multirow{1}{*}{\rmnum{2}} &CDIKTNet &10& \textbf{83.60} &\textbf{86.24} &\textbf{90.53}  &\textbf{92.14} & \textbf{94.33} &\textbf{95.32} &\textbf{97.13}  &\textbf{97.60} &\textbf{91.25} &\textbf{83.24} &\textbf{95.00} &\textbf{89.52}  &\textbf{97.50} &\textbf{93.59} &\textbf{97.50} &\textbf{95.80}\\\hline
\end{tabular}%
}
\footnote{3}{\footnotesize“--” denotes the scenario where supervised methods are applied under limited paired relationships. \qquad\qquad\qquad\qquad\qquad\qquad\qquad\qquad\qquad\qquad\qquad}
\label{tab:university-compare-sues}
\end{table*}

\begin{table}[t]
\centering
\tiny
\caption{Comparisons between existing state-of-the-art methods and CDIKTNet  on University-1652 with 2\% and 10\% of paired data.}
\resizebox{\columnwidth}{!}{ 
\setlength{\tabcolsep}{3pt}
\begin{tabular}{ccccccc}\hline
\multicolumn{7}{c}{University-1652}  \\ \hline
\multirow{2}{*}{Setting}&\multirow{2}{*}{Model} & \multirow{2}{*}{\makecell[c]{GT Ratio\\(\%)}} & \multicolumn{2}{c}{Drone$\rightarrow$Satellite} & \multicolumn{2}{c}{Satellite$\rightarrow$Drone} \\ \cline{4-7} 
&&  & R@1 & AP   & R@1 & AP\\ \hline
  \multirow{3}{*}{-}&Sample4Geo\cite{deuser2023sample4geo}  &2       & 62.41      &66.85       & 84.45     &60.52 \\\
      &CAMP\cite{wu2024camp}                &2  &{62.53}   &{67.01}    &{83.31}     &{59.26} \\
  &DAC\cite{xia2024enhancing}          &2  &{63.68} &{68.16} &{87.59}&{61.26} \\ \hline

\multirow{1}{*}{\rmnum{2}}&CDIKTNet    &2 &\textbf{85.63}       &\textbf{87.75} &\textbf{90.44}     & \textbf{82.24}\\\hline
  \multirow{3}{*}{-} &CAMP\cite{wu2024camp}                &10  & 83.64      &86.27       & 92.44     & 82.27 \\
 &Sample4Geo\cite{deuser2023sample4geo}  &10       & 84.51      &86.94       & 93.01    & {83.53} \\
  &DAC\cite{xia2024enhancing}          &10  &{84.64} &{87.07} &{93.15}&{83.03} \\\hline
\multirow{1}{*}{\rmnum{2}}&CDIKTNet    &10 &\textbf{88.02}       &\textbf{89.84} &\textbf{93.87}     &\textbf{86.48}\\\hline
\end{tabular}}
\label{tab:university-compare-u1652}

\end{table}
\textbf{(\rmnum{2}) Few-shot Initialization}: To evaluate the effectiveness of CDIKTNet in leveraging unlabeled data for learning, we initialize CDIKTNet with only 2\% and 10\% of the paired data. The remaining training data are entirely unpaired. {\color{black} As shown in Tables \ref{tab:university} and \ref{tab:SUES-200}, CDIKTNet with only 2\% of the paired data achieves performance comparable to or even surpassing that of IFSs \cite{ge2024multibranch}.} With 10\% of paired data, CDIKTNet achieves further improvements, surpassing most fully supervised methods and achieving performance comparable to MCCG \cite{shen2023mccg}. Notably, as shown in Table~\ref{tab:university-seus}, CDIKTNet in this setting trained with only 10\% of the paired data exhibits even stronger generalization than these fully supervised methods. 

Furthermore, compared with state-of-the-art unsupervised methods, including EM-CVGL \cite{10589921}, AnyLoc \cite{keetha2023anyloc}, SegVLAD \cite{garg2024revisit}, and Li \cite{11010141}, as illustrated in Tables \ref{tab:university}, \ref{tab:SUES-200}, \ref{tab:university1652} and \ref{tab:sues-200}, CDIKTNet in this setting significantly outperforms them in all metrics on both datasets, despite relying on only a small proportion of paired data. In terms of generalization, CDIKTNet still significantly surpasses these methods, as reported in Tables \ref{tab:university-seus} and \ref{tab:sues-20012}. This underscores the significance of employing a limited quantity (e.g., 2\%) of paired data for initialization, as it serves to steer the exploration toward revealing deeper cross-view representations from unpaired data.

{\color{black}
\label{Cross-domain Initialization}\textbf{(\rmnum{3}) Cross-domain Initialization}: 
A novel experimental setting is introduced for cross-domain validation. CDIKTNet is first initialized using 2\% paired data from one dataset and then trained with the remaining unpaired samples in that source domain. The trained model is subsequently transferred to another dataset representing a distinct target domain for further adaptation without requiring additional pairing relationships. Notably, no paired supervision is used in the target domain during this transfer phase. Therefore, the GT ratio is marked as 2/0 in this paper, which denotes 2\% paired supervision in the source domain and 0\% paired supervision in the target domain, rather than 0\% paired supervision throughout the entire training. Since limited paired supervision is used for source-domain initialization, this setting is not strictly equivalent to genuinely unsupervised methods, but should instead be viewed as a special setting that lies between supervised and unsupervised cross-domain learning. Despite this distinction, the setting remains practically valuable, as it reflects a realistic transfer scenario and allows us to evaluate the cross-domain adaptation capability of the proposed method under zero paired supervision in the target domain.}

In setting~\rmnum{3}~of Tables \ref{tab:university}, and \ref{tab:university1652},  CDIKTNet is first initialized by using 2\% paired data on SUES-200 dataset and then continue training with the remaining unpaired data. The resulting model has corresponded to setting~\rmnum{2}~with GT Ratio 2\% in Table \ref{tab:SUES-200}. This model is subsequently used as a cross-domain initialization and transferred to the other dataset University-1652 for further training, where no paired data are required in the following process. Similarly, in setting~\rmnum{3}~of Tables \ref{tab:SUES-200} and \ref{tab:sues-200}, CDIKTNet is initialized using 2\% paired data on the University-1652 dataset and continue training with its remaining unpaired data. This trained model has reported in setting~\rmnum{2}~with GT Ratio 2\% in Table \ref{tab:university}. This model is then used as a cross-domain initialization on the SUES-200 dataset, where the transfer stage also relies entirely on unpaired data on this new dataset. Furthermore, the results in setting~\rmnum{3}~of Tables \ref{tab:university-seus} and \ref{tab:sues-20012} are obtained by directly evaluating the models from setting~\rmnum{3}~of Tables \ref{tab:university} and \ref{tab:university1652} on the SUES-200 dataset, without requiring any additional fine-tuning.

As shown in Tables \ref{tab:university}, \ref{tab:SUES-200}, \ref{tab:university1652} and \ref{tab:sues-200}, after training in one domain, CDIKTNet can be directly transferred to another domain. It outperforms state-of-the-art unsupervised methods in all metrics and achieves performance comparable to several fully supervised methods. Moreover, as reported in Tables \ref{tab:university-seus} and \ref{tab:sues-20012}, CDIKTNet demonstrates remarkable generalization performance. It significantly outperforms these state-of-the-art unsupervised methods. Notably, as shown in Table \ref{tab:sues-200}, the experimental results in this setting demonstrate superior performance compared with the few-shot initialization setting, which has utilized 2\% of paired data in the current domain. This indicates that strong domain adaptability enables the cross-domain initialization setting to outperform limited paired supervision under some cases.

\begin{table}[t]
\centering
\caption{Influence of each component on performance of CDIKTNet.}
\tiny
\resizebox{\columnwidth}{!}{ 
\setlength{\tabcolsep}{3pt}
\begin{tabular}{lcccccccc}\hline
\multicolumn{8}{c}{University-1652}  \\ \hline
& \multirow{2}{*}{SIFL} & \multirow{2}{*}{SIEL} & \multirow{2}{*}{CDTS} & \multicolumn{2}{c}{Drone$\rightarrow$Satellite} & \multicolumn{2}{c}{Satellite$\rightarrow$Drone} \\ \cline{5-8}
&&&& R@1 & AP & R@1 & AP \\ \hline
\multirow{5}{*}{\rotatebox{90}{2\%}} &  &  &  & 58.19 & 62.69 & 82.45 & 56.95 \\
 & $\checkmark$ &  &  &{59.00} & {63.55} & {81.16} & {55.03} \\
  &  & $\checkmark$ & & {59.73} & {64.16} &{80.03}  & {57.35} \\
  & $\checkmark$ &$\checkmark$  &  &{63.93} &{68.19} &{83.45} & {61.35} \\
 & $\checkmark$ & $\checkmark$ &$\checkmark$  &\textbf{85.63} &\textbf{87.75} &\textbf{90.44} &\textbf{82.24} \\ \hline
 \multirow{4}{*}{\rotatebox{90}{100\%}} &  &  &  & 92.80 & 94.00 & 95.01 & 91.83 \\
 & $\checkmark$ &  &  & {94.02} & {94.96} & {95.72} & {93.16} \\
  &  &$\checkmark$  &  & {93.26} & {94.36} & {95.58} & {92.15} \\
 & $\checkmark$ & $\checkmark$ & &\textbf{95.02}       &\textbf{95.79} &\textbf{96.15}     &\textbf{93.93}\\ \hline
\end{tabular}}
\label{tab:university-ab}
\end{table}

\remark{ The (\rmnum{2}) few-shot initialization setting and (\rmnum{3}) cross-domain initialization} setting are two typical scenarios of limited supervision. The former targets the current domain by using a limited amount of paired data for initialization, which is a process of supervised learning. Subsequently, unpaired data is employed for unsupervised learning. The latter addresses the issue of a complete lack of paired data in a new domain. First, it performs initialization through supervised learning in an existing domain using a limited amount of paired data. Then, it uses the remaining unpaired data for unsupervised learning. The obtained model is directly applied to unsupervised learning in the new domain, and has achieved remarkable results in the DVGL task in the new domain.

Furthermore, as shown in Tables \ref{tab:university-compare-sues} and \ref{tab:university-compare-u1652}, existing supervised methods strictly rely on paired data as positive samples and cannot use unpaired data. In contrast, CDIKTNet can effectively exploit a large amount of unpaired cross-view information to achieve better performance under such limited conditions. Additionally, CDIKTNet also achieves competitive performance with 100\% of paired data equivalent to fully supervised learning. This highlights its excellent adaptability and generalization in different supervision levels.

\begin{figure}[h]
  \centering
  \includegraphics[width=3.5in]{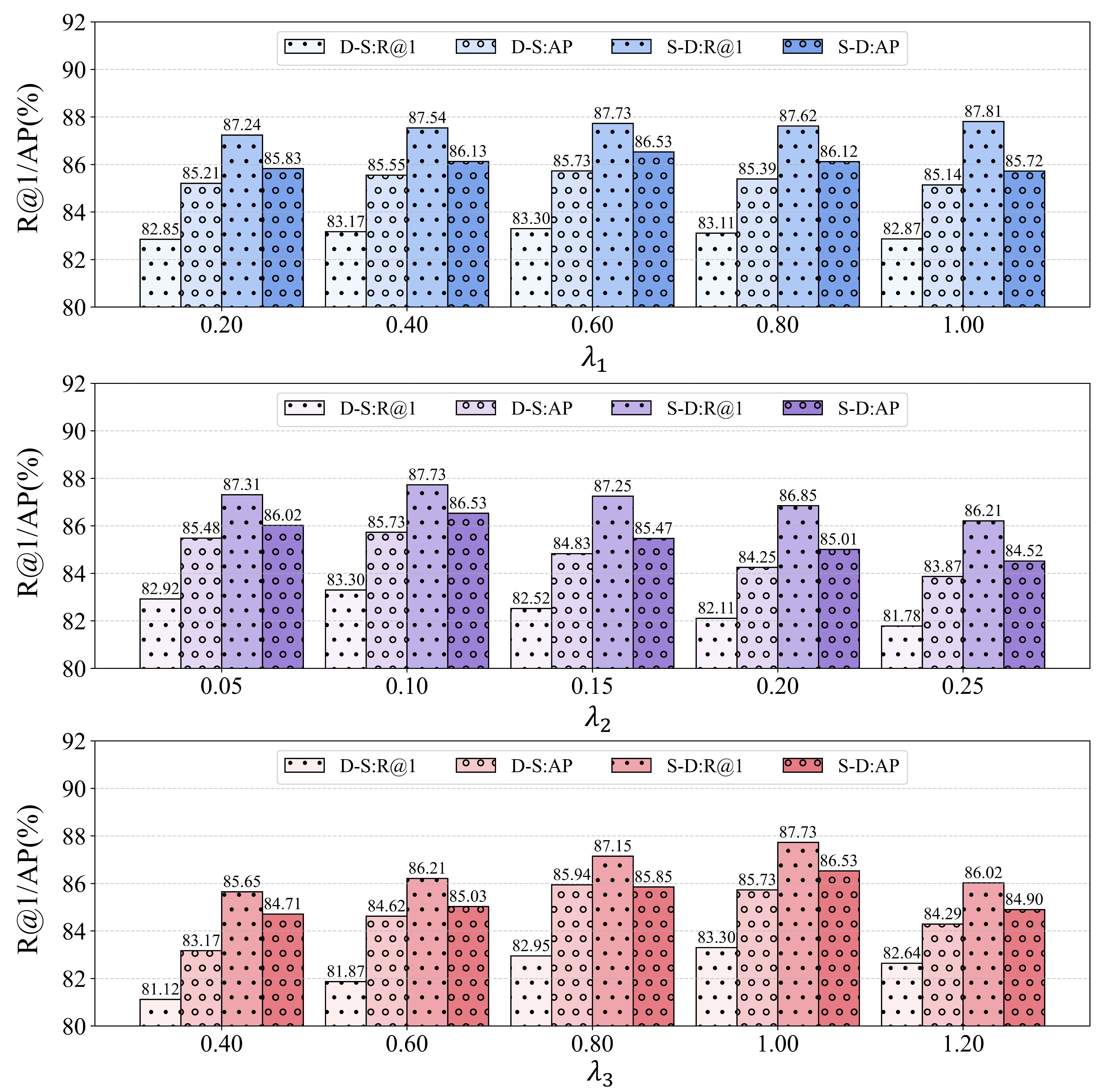}
  \caption{\textbf{Performance Influence of $\lambda_1$, $\lambda_2$ and $\lambda_3$}. Influence of different $\lambda_1$, $\lambda_2$ and $\lambda_3$ values on University-1652 in setting \rmnum{3}.}
  \label{fig7}
  \end{figure}

\begin{figure}[h]
  \centering
  \includegraphics[width=3.5in]{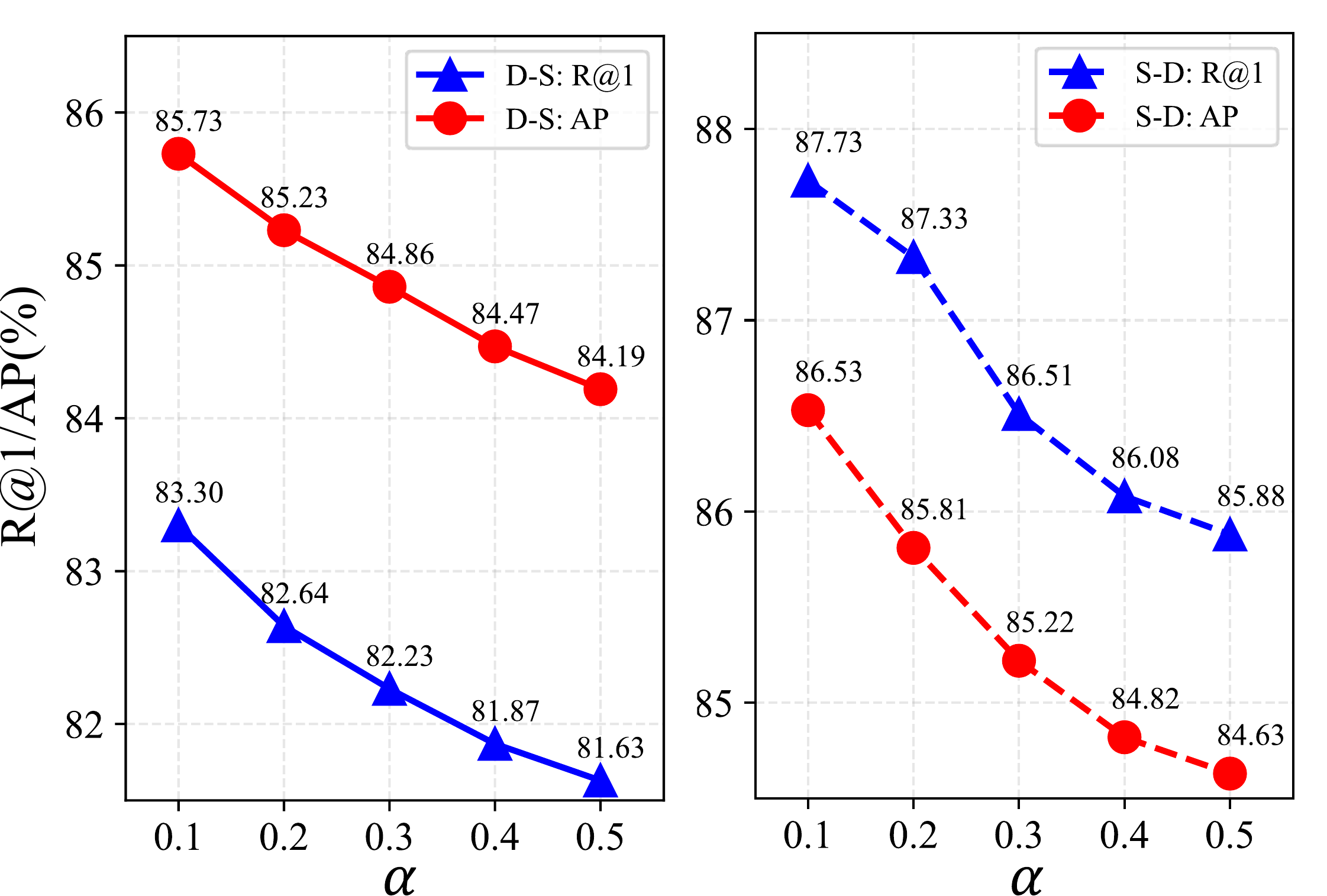}
  \caption{\textbf{Performance Influence of $\alpha$}. Influence of different $\alpha$ values on University-1652 in setting \rmnum{3}.}
  \label{fig8}
  \end{figure}
  
\subsection{Ablation Studies}
\textbf{Effectiveness of CDIS}. As shown in Table \ref{tab:university-ab}, both the SIFL block and SIEL module contribute significantly to performance improvements with both 2\% and 100\% of paired data. When used individually, each component increases R@1 and AP, which demonstrates their individual effectiveness. Notably, their joint optimization yields further improvement, which validates the design of CDIS. Specifically, SIFL enhances geometric invariance at different viewpoints by modeling structural priors, while SIEL improves spatial invariance by incorporating scale-aware contextual cues. These two components complement each other to enable CDIKTNet to construct a unified and robust cross-view shared feature space.

\textbf{Effectiveness of CDTS}. With 2\% of paired data, the introduction of CDTS leads to an increase in R@1 from 63.93\% to 85.63\% and AP from 68.19\% to 87.75\% in the drone$\rightarrow$satellite scenario, while also improving satellite$\rightarrow$drone R@1 from 83.45\% to 90.44\% and AP from 61.35\% to 82.24\%. These gains clearly show the effectiveness of CDTS. Using the invariant knowledge learned by CDIS, CDTS enables a more effective mining of cross-view correlations from unpaired data. It makes CDIKTNet retain strong performance in conditions with scarce paired annotations.

\begin{figure}[h]
  \centering
  \includegraphics[width=3.5in]{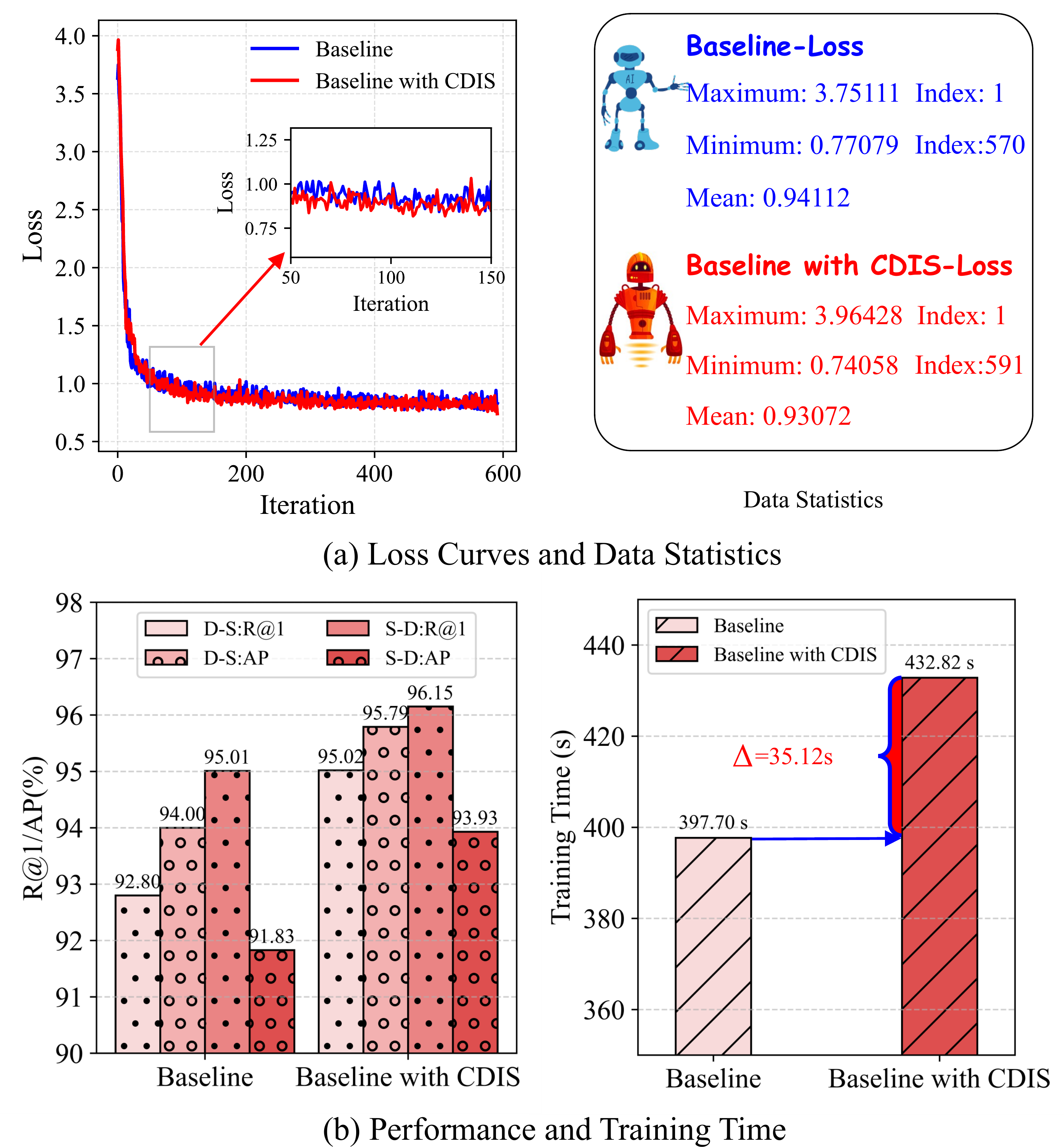}
  \caption{\textbf{Influence of CDIS on Convergence, Performance, and Training Efficiency.} (a) Loss curves and statistical summaries for baseline vs. baseline with CDIS on University-1652 in setting \rmnum{1}. (b) Comparison of performance and training time between baseline and baseline with CDIS on University-1652 in setting \rmnum{1}.}
  \label{fig9}
  \end{figure}

{\color{black}\subsection{Further Analysis}
\textbf{Hyper-parameters Analysis of Loss Coefficients.}
We further investigate the influence of the loss coefficients $\lambda_1$, $\lambda_2$, and $\lambda_3$ in Eq.~~(\ref{eq7}) on the University-1652 dataset. As shown in Fig.~\ref{fig7}, the performance influence of all three hyper-parameters exhibit bounded fluctuations within reasonable ranges. It achieves consistently robust performance for $\lambda_1 \in [0.4,0.8]$, and values outside this range diminish the effectiveness. Then the peak performance occurs near $\lambda_2 = 0.10$, with marginal degradation observed at both lower and higher values. Furthermore, it exhibits a gradual improvement as its weight increases from 0.40 to 1.00 for $\lambda_3$, in which the best results are obtained when $\lambda_3 = 1.00$. Synthesizing these observations, the highest overall performance is attained at  $\lambda_1 = 0.60$, $\lambda_2 = 0.10$, and $\lambda_3 = 1.00$. Consequently, these coefficient values are adopted for all subsequent experiments.
}

{\color{black}
\textbf{Hyper-parameters Analysis of $\alpha$.}
To evaluate the performance influence of $\alpha$, quantitative comparisons are given in Fig.~\ref{fig8}. {\color{black}As shown in Fig.~\ref{fig8}, increasing the momentum factor $\alpha$ leads to degraded performance in both drone$\rightarrow$satellite and satellite$\rightarrow$drone scenarios.} A larger $\alpha$ slows the update of the cluster memories and reduces the adaptability  to cross-view variations. In contrast, CDTS requires rapid acquisition of cross-view cues, and a smaller $\alpha$ assigns greater weight to new features, which allows CDTS to achieve faster adaptation.}

{\color{black}\textbf{Influence of CDIS on Convergence, Performance, and Training Efficiency.}
To further evaluate the influence of CDIS on convergence, the training behaviors of the baseline and baseline with CDIS are compared on the University-1652 dataset in setting \rmnum{1}. As shown in Fig. \ref{fig9}(a), it presents the loss curves and their statistical values. The baseline and baseline with CDIS exhibit similar descending trends and stable convergence regions. The baseline with CDIS begins with a slightly higher loss but reaches a lower final value, which demonstrates that the optimization process remains stable and that CDIS guides the model toward a better solution. Fig.~\ref{fig9}(b) reports the performance and the training time. The CDIS improves the performance in both the drone $\rightarrow$ satellite and satellite $\rightarrow$ drone scenarios compared to the baseline, and the additional training cost is small at about 35 seconds. Empirical evidence confirms that CDIS simultaneously enhances structural and spatial invariance, which maintains rapid convergence rates and incurs only marginal computational overhead, ultimately achieving superior performance.
}

\begin{figure*}[t]
  \centering
  \includegraphics[width=7.2in]{4.pdf}
  \caption{\textbf{Similarity Distribution and Feature Space Visualization on University-1652 }. (a–e) show the similarity distribution visualization. The red-shaded area indicates the overlapping region between the positive and negative distributions. (f-j) show the distribution of feature embeddings in the 2D shared feature space, where circles and triangles in different colors denote drone-view and satellite-view. A total of 20 localizations are  randomly selected from the test set.}
  \label{fig4}
  \end{figure*}

\begin{table}[t]
  \caption{Comparisons between existing state-of-the-art methods and CDIKTNet on DenseUAV}
  \label{tab:DenseUAV}
  \footnotesize
\centering
\setlength{\tabcolsep}{3pt}
\resizebox{\columnwidth}{!}{ 
  \begin{tabular}{ccccccccccc}
  \hline
   &  \multicolumn{10}{c}{DenseUAV} \\ \cline{1-11}
   &    && \multicolumn{2}{c}{All height}& \multicolumn{2}{c}{80m} &\multicolumn{2}{c}{90m}& \multicolumn{2}{c}{100m}  \\ \cline{4-11}
\multirow{-2}{*}{Setting} & \multirow{-2}{*}{Model} & \multirow{-2}{*}{\makecell[c]{GT Ratio\\(\%)}} & {R@1} & {R@5} & {R@1} & {R@5} & {R@1} & {R@5} & {R@1} & {R@5}\\ \hline
\multirow{4}{*}{\rmnum{1}} & ConvNext-Tiny\cite{DenseUAV}       & 100  & 60.23 & 81.94 & -  & - & -  & -& -  & -\\
         &Sample4Geo\cite{deuser2023sample4geo}              & 100& 80.57 &96.53  &80.57&96.53 &{86.87}&98.71 &{91.38}& {99.74} \\
         &DAC\cite{xia2024enhancing}             &100 & 84.47 & 96.53 &80.92 &97.04  &85.32 &98.71 &85.46 &99.10\\ 
         &CDIKTNet    &100 & \textbf{85.50} &  \textbf{97.53} & \textbf{83.14} & \textbf{97.55}  & \textbf{86.36} & \textbf{99.23} & \textbf{87.39} & \textbf{99.49}  \\\hline
\multirow{2}{*}{\rmnum{2}} & CDIKTNet &{10} &{51.57} &{79.37} &{49.93} &{83.91} &{52.90} &{84.93} &{52.12}   &{88.28}\\
&CDIKTNet     & 2        &{39.76} & {72.11} &{39.00} &{75.55} &{39.64} &{78.77} &{39.90}  &{80.31}  \\\hline
\multirow{1}{*}{{UM}}&EM-CVGL\cite{10589921}  &0  & 18.15  & 50.97       & 18.02      &50.36   & 17.76  & 50.19       & 17.50      & 49.98    \\\hline
\multirow{1}{*}{\rmnum{3}}&CDIKTNet &2/0  &\textbf{22.57}   &\textbf{40.24}   &\textbf{17.25} &\textbf{40.15}   &\textbf{23.04}   &\textbf{48.39}   &\textbf{25.48}    &\textbf{52.38}\\\hline
\end{tabular}}
\end{table}

{\color{black}
\textbf{Effectiveness and Limitations Discussions.} 
 To further validate the effectiveness of CDIKTNet,  a more challenging dataset DenseUAV \cite{DenseUAV} is introduced, which focuses on low-altitude drone self-positioning. This dataset includes over 27,000 drone$\rightarrow$satellite images collected from 14 university campuses. DenseUAV presents significant challenges such as cross-view misalignment and spatial-temporal variations, with multi-scale satellite imagery and diverse temporal conditions. As shown in Table \ref{tab:DenseUAV}, in setting \rmnum{1}, CDIKTNet surpasses most supervised methods despite the dataset's complexity. In setting \rmnum{2}, by trading off a large portion of paired data, CDIKTNet also achieves substantial performance gains over EM-CVGL. Moreover, in setting \rmnum{3}, even when initialized via cross-domain transfer from University-1652, CDIKTNet maintains superiority over EM-CVGL. However, in settings \rmnum{2}~and \rmnum{3}, CDIKTNet performs suboptimally compared to current supervised methods. The main challenge lies in the inherent confusion within the DenseUAV dataset. The DenseUAV dataset's intrinsic cross-view misalignment and spatio-temporal variations become more pronounced under limited paired-data initialization. Additionally, effective transfer requires the model to embed domain-invariant features during source-domain pretraining, which remains challenging without strong prior knowledge. The initialization phase demands robust prior knowledge to enhance cross-view alignment, as weaker priors lead to suboptimal performance.  In future work, we will explore the methods that do not rely on such priors.
}

 \begin{figure}[h]
  \centering
  \includegraphics[width=3.5in]{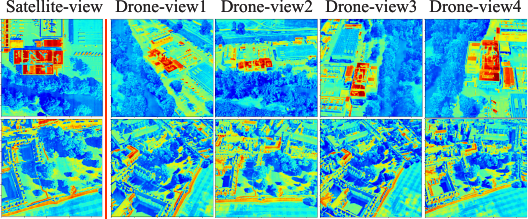}
  \caption{\textbf{Attention Heatmap Visualization}. The attention responses generated by the proposed SIEL module are visualized. Red regions correspond to stronger attention responses, whereas blue regions denote weaker responses.}
  \label{fig5}
  \end{figure}
  \begin{figure}[h]
  \centering
  \includegraphics[width=3.4in]{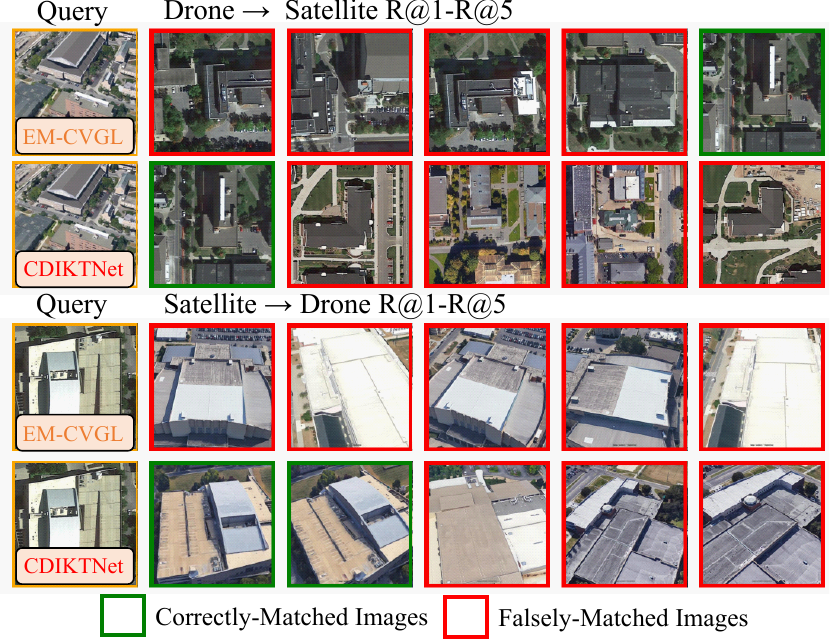}
  \caption{\textbf{Retrieval Visualization on University-1652.} The green box indicates a
correctly-matched image, and the red box indicates a falsely-matched image.}
  \label{fig6}
  \end{figure}
  
\subsection{Visualization}
As shown in Fig. \ref{fig4}, to further demonstrate the effectiveness of CDIKTNet, we visualize the similarity distribution using t-SNE \cite{van2008visualizing} and feature space projection.

\textbf{Similarity Distribution Visualization}. As shown in Fig. \ref{fig4}(a–b), Sample4Geo and DAC achieve overlap areas of 0.1232 and 0.1124 under limited supervision, due to their inability to exploit unpaired data. Fig. \ref{fig4}(c) shows that EM-CVGL \cite{10589921} achieves an overlap of 0.1158. Although it performs slightly better, feature confusion remains an issue. In contrast, CDIKTNet in Fig. \ref{fig4}(d) achieves a lower overlap of 0.0661 with only 2\% of paired data. This confirms its strong ability to align and separate cross-view features under limited supervision. Notably, in Fig. \ref{fig4}(e), CDIKTNet is trained in the cross-domain initialization setting described in Subsection \ref{Cross-domain Initialization}. It still achieves an overlap of 0.0808 to outperform all methods. This shows its effectiveness in exploiting unpaired data and transferring knowledge from an external source domain.

\textbf{Feature Space Visualization}. In the feature space, features from the same location should cluster, while different locations should be well separated. As shown in Fig. \ref{fig4}(f–h), the compared methods show noticeable cross-view confusion (red circles) and dispersed feature distributions. In contrast, CDIKTNet (Fig. \ref{fig4}(i–j)) produces more compact and structured feature distributions in both scenarios. Although the 0\% of paired data in cross-domain initialization setting shows minor confusion, the features remain more compact and more evenly aligned. This presents the ability of CDIKTNet to learn cross-view information.

\textbf{Attention Heatmap Visualization}.
As a further demonstration of the effectiveness of SIEL, we visualize its attention heat maps in Fig. \ref{fig5}. Despite the large spatial differences between satellite-view and drone-view images and the feature distribution shifts caused by varying drone viewpoints, SIEL consistently highlights semantically meaningful regions in both views.

\textbf{Retrieval Visualization}. To visualize the effectiveness of CDIKTNet in cross-domain initialization setting, Fig. \ref{fig6} shows retrieval results in both drone$\rightarrow$satellite and satellite$\rightarrow$drone scenarios. CDIKTNet retrieves more correct matches than EM-CVGL \cite{10589921}, which demonstrates superior cross-view alignment and robustness to viewpoint variations.

\section{Conclusion}\label{conclusion}
This paper addresses a critical and highly challenging problem in the drone-view geo-localization (DVGL) task with minimal or even no paired data to reduce dependence on expensive cross-view annotations. To this end, we propose a novel cross-domain invariant knowledge transfer network (CDIKTNet), which learns the spatial and structural invariance of cross-view features while constructing a robust invariant knowledge representation space. Meanwhile, it utilizes this shared space as an optimization anchor for unsupervised training, which effectively learns latent cross-view information from unpaired data. It significantly alleviates the reliance on paired data. More importantly, CDIKTNet can be applied under any level of supervision and can be generalized to new environments without requiring their prior pairing relationships. Extensive experiments demonstrate that CDIKTNet achieves competitive performance under full supervision, compared with existing supervised methods. Furthermore, when initialized with a small amount of paired data in the current domain or transferred to the other domain in the proposed cross-domain initialization setting, CDIKTNet outperforms state-of-the-art unsupervised methods and even surpasses some fully supervised methods. This advancement pushes DVGL closer to real-world deployment.

\bibliographystyle{IEEEtran}
\footnotesize
\bibliography{main}

@String(AAAI = {AAAI})

@inproceedings{deuser2023sample4geo,
  title={Sample4{G}eo: {H}ard negative sampling for cross-view geo-localisation},
  author={Deuser, Fabian and Habel, Konrad and Oswald, Norbert},
  booktitle = {Proc. IEEE Conf. Comput. Vis. Pattern Recognit.},
  year= {2023},
  pages= {16847-16856}
}

@article{shen2023mccg,
  title={{MCCG}: {A} convNeXt-based multiple-classifier method for cross-view geo-localization},
  author={Shen, Tianrui and Wei, Yingmei and Kang, Lai and Wan, Shanshan and Yang, Yee-Hong},
  journal={{IEEE} Trans. Circuits Syst. Video Technol.},
  volume={34},
  number={3},
  pages={1456-1468},
  year={2023},
  publisher={IEEE}
}

@article{zhao2024transfg,
  title={{T}rans{FG}: {A} cross-view geo-localization of satellite and {UAV}s imagery pipeline using transformer-based feature aggregation and gradient guidance},
  author={Zhao, Hu and Ren, Keyan and Yue, Tianyi and Zhang, Chun and Yuan, Shuai},
  journal={{IEEE} Trans. Geosci. Remote Sens.},
  volume={62},
  number={},
  pages={1-12},
  year={2024},
  publisher={IEEE}
}

@article{ge2024multibranch,
  title={Multibranch joint representation learning based on information fusion strategy for cross-view geo-localization},
  author={Ge, Fawei and Zhang, Yunzhou and Liu, Yixiu and Wang, Guiyuan and Coleman, Sonya and Kerr, Dermot and Wang, Li},
  journal={{IEEE} Trans. Geosci. Remote Sens.},
  volume={62},
  pages={1-16},
  year={2024},
  publisher={IEEE}
}

@article{xia2024enhancing,
  title={Enhancing cross-view geo-localization with domain alignment and scene consistency},
  author={Xia, Panwang and Wan, Yi and Zheng, Zhi and Zhang, Yongjun and Deng, Jiwei},
  journal={{IEEE} Trans. Circuits Syst. Video Technol.},
  year={2024},
  volume={},
  number={},
  pages={1-12},
  publisher={IEEE}
}

@article{tian2021uav,
  title={{UAV}-satellite view synthesis for cross-view geo-localization},
  author={Tian, Xiaoyang and Shao, Jie and Ouyang, Deqiang and Shen, Heng Tao},
  journal={{IEEE} Trans. Circuits Syst. Video Technol.},
  volume={32},
  number={7},
  pages={4804-4815},
  year={2021},
  publisher={IEEE}
}

@inproceedings{liu2022convnet,
  title={A convNet for the 2020s},
  author={Liu, Zhuang and Mao, Hanzi and Wu, Chao-Yuan and Feichtenhofer, Christoph and Darrell, Trevor and Xie, Saining},
  booktitle = {Proc. IEEE Conf. Comput. Vis. Pattern Recognit.},
  year= {2022},
  pages = {11976-11986}

}

@article{du2024ccr,
  title={{CCR}: A counterfactual causal reasoning-based method for cross-view geo-localization},
  author={Du, Haolin and He, Jingfei and Zhao, Yuanqing},
  journal={{IEEE} Trans. Circuits Syst. Video Technol.},
  year={2024},
  volume={34},
  number={11},
  pages={11630-11643},
}

@article{wu2024camp,
  title={{CAMP}: {A}cross-view geo-localization method using contrastive attributes mining and position-aware partitioning},
  author={Wu, Qiong and Wan, Yi and Zheng, Zhi and Zhang, Yongjun and Wang, Guangshuai and Zhao, Zhenyang},
  journal={{IEEE} Trans. Geosci. Remote Sens.},
  year={2024},
  volume={62},
  number={},
  pages={1-14},
}

@article{lv2024direction,
  title={Direction-guided multi-scale feature fusion network for geo-localization},
  author={Lv, Hongxiang and Zhu, Hai and Zhu, Runzhe and Wu, Fei and Wang, Chunyuan and Cai, Meiyu and Zhang, Kaiyu},
  journal={{IEEE} Trans. Geosci. Remote Sens.},
  year={2024},
  volume={62},
  number={},
  pages={1-13},
  publisher={IEEE}
}

@article{oord2018representation,
  title={Representation learning with contrastive predictive coding},
  author={Oord, Aaron van den and Li, Yazhe and Vinyals, Oriol},
  journal={arXiv preprint arXiv:1807.03748},
  year={2018}
}

@article{wang2024multiple,
  title={Multiple-environment self-adaptive network for aerial-view geo-localization},
  author={Wang, Tingyu and Zheng, Zhedong and Sun, Yaoqi and Yan, Chenggang and Yang, Yi and Chua, Tat-Seng},
  journal={Pattern Recognit.},
  volume={152},
  pages={110363},
  year={2024},
  publisher={Elsevier}
}

@article{zhu2023sues,
  title={{SUES}-200: {A} multi-height multi-scene cross-view image benchmark across drone and satellite},
  author={Zhu, Runzhe and Yin, Ling and Yang, Mingze and Wu, Fei and Yang, Yuncheng and Hu, Wenbo},
  journal={{IEEE} Trans. Circuits Syst. Video Technol.},
  volume={33},
  number={9},
  pages={4825-4839},
  year={2023},
  publisher={IEEE}
}

@inproceedings{zheng2020university,
author = {Zheng, Zhedong and Wei, Yunchao and Yang, Yi},
title = {University-1652: {A} Multi-view Multi-source Benchmark for Drone-based Geo-localization},
year = {2020},
booktitle = {Proc. ACM Int. Conf. Multimedia},
pages = {1395–1403},
}

@article{van2008visualizing,
  title={{V}isualizing data using {t-SNE}.},
  author={Van der Maaten, Laurens and Hinton, Geoffrey},
  journal={J. Mach. Learn. Res},
  volume={9},
  number={11},
  pages={2579-2605},
  year={2008}
}

@article{cui2021cross,
  title={Cross-modality image matching network With modality-invariant feature representation for airborne-ground thermal infrared and visible datasets},
  author={Cui, Song and Ma, Ailong and Wan, Yuting and Zhong, Yanfei and Luo, Bin and Xu, Miaozhong},
  journal={{IEEE} Trans. Geosci. Remote Sens.},
  volume={60},
  pages={1-14},
  year={2021},
  publisher={IEEE}
}

@article{chen2024multi,
  title={Multilevel Embedding and Alignment Network With Consistency and Invariance Learning for Cross-View Geo-Localization}, 
  author={Chen, Zhongwei and Yang, Zhao-Xu and Rong, Hai-Jun},
  journal={{IEEE} Trans. Geosci. Remote Sens.}, 
  year={2025},
  volume={63},
  number={},
  pages={1-15},
}

@article{DenseUAV,
  author={Dai, Ming and Zheng, Enhui and Feng, Zhenhua and Qi, Lei and Zhuang, Jiedong and Yang, Wankou},
  journal={{IEEE} Trans. Image Process.},
  title={Vision-Based UAV Self-Positioning in Low-Altitude Urban Environments},
  year={2024},
  volume={33},
  number={},
  pages={493-508}
}

@inproceedings{dai2022cluster,
  title={Cluster contrast for unsupervised person re-identification},
  author={Dai, Zuozhuo and Wang, Guangyuan and Yuan, Weihao and Zhu, Siyu and Tan, Ping},
  booktitle={Proc. Asian Conf. Comput. Vis.},
  pages={1142-1160},
  year={2022}
}

@article{goodfellow2014generative,
  title={Generative Adversarial Nets},
  author={Goodfellow, Ian and Pouget-Abadie, Jean and Mirza, Mehdi and Xu, Bing and Warde-Farley, David and Ozair, Sherjil and Courville, Aaron and Bengio, Yoshua},
  journal={in Proc. Adv. Neural Inf. Process. Syst.},
  volume={27},
  pages={2672–2680},
  year={2014},
}

@inproceedings{ester1996density,
  title={A density-based algorithm for discovering clusters in large spatial databases with noise},
  author={Ester, Martin and Kriegel, Hans-Peter and Sander, J{\"o}rg and Xu, Xiaowei and others},
  booktitle={Proc. ACM SIGKDD Int. Conf. Knowl. Discov. Data Min.},
  volume={96},
  number={34},
  pages={226-231},
  year={1996}
}

@inproceedings{adca,
  title={Augmented Dual-Contrastive Aggregation Learning for Unsupervised Visible-Infrared Person Re-Identification},
  author={Yang, Bin and Ye, Mang and Chen, Jun and Wu, Zesen},
  pages = {2843–2851},
  booktitle = {ACM MM},
  year={2022}
}

@inproceedings{zhang2023cross,
  title={{C}ross-view geo-localization via learning disentangled geometric layout correspondence},
  author={Zhang, Xiaohan and Li, Xingyu and Sultani, Waqas and Zhou, Yi and Wshah, Safwan},
  booktitle={Proc. AAAI Conf. Artif. Intell.},
  volume={37},
  number={3},
  pages={3480-3488},
  year={2023}
}

@inproceedings{yang2024shallow,
  title={Shallow-Deep Collaborative Learning for Unsupervised Visible-Infrared Person Re-Identification},
  author={Yang, Bin and Chen, Jun and Ye, Mang},
  booktitle={Proc. IEEE Conf. Comput. Vis. Pattern Recognit.},
  pages={16870-16879},
  year={2024}
}

@InProceedings{Cai_2019_ICCV,
author = {Cai, Sudong and Guo, Yulan and Khan, Salman and Hu, Jiwei and Wen, Gongjian},
title = {Ground-to-Aerial Image Geo-Localization With a Hard Exemplar Reweighting Triplet Loss},
booktitle = {Proc. IEEE Int. Conf. Comput. Vis.},
year = {2019},
pages= {8391-8400}
}

@inproceedings{Hu,
author = {Hu, Wenmiao and Zhang, Yichen and Liang, Yuxuan and Yin, Yifang and Georgescu, Andrei and Tran, An and Kruppa, Hannes and Ng, See-Kiong and Zimmermann, Roger},
title = {Beyond Geo-localization: Fine-grained Orientation of Street-view Images by Cross-view Matching with Satellite Imagery},
year = {2022},
booktitle = {Proc. ACM Int. Conf. Multimedia},
pages = {6155–6164},
numpages = {10},
}

@InProceedings{Liu_2019_CVPR,
author = {Liu, Liu and Li, Hongdong},
title = {Lending Orientation to Neural Networks for Cross-View Geo-Localization},
booktitle = {Proc. IEEE Conf. Comput. Vis. Pattern Recognit.},
pages = {5624–5633},
year = {2019}
}

@InProceedings{Shi_Yu_Liu_Zhang_Li_2020, 
title={Optimal Feature Transport for Cross-View Image Geo-Localization}, 
volume={34},
booktitle={Proc. AAAI Conf. Artif. Intell. }, 
author={Shi, Yujiao and Yu, Xin and Liu, Liu and Zhang, Tong and Li, Hongdong}, 
year={2020}, 
pages={11990-11997} }

@inproceedings{yang2022augmented,
  title={Augmented dual-contrastive aggregation learning for unsupervised visible-infrared person re-identification},
  author={Yang, Bin and Ye, Mang and Chen, Jun and Wu, Zesen},
  booktitle={Proc. ACM Int. Conf. Multimedia},
  pages={2843-2851},
  year={2022}
}

@InProceedings{Liu_Multi,
    author    = {Liu, Shuai and Li, Xin and Lu, Huchuan and He, You},
    title     = {Multi-Object Tracking Meets Moving UAV},
    booktitle = {Proc. IEEE Conf. Comput. Vis. Pattern Recognit.},
    year      = {2022},
    pages     = {8876-8885}
}

@InProceedings{Li_PVT++,
    author    = {Li, Bowen and Huang, Ziyuan and Ye, Junjie and Li, Yiming and Scherer, Sebastian and Zhao, Hang and Fu, Changhong},
    title     = {PVT++: A Simple End-to-End Latency-Aware Visual Tracking Framework},
    booktitle = {Proc. IEEE Int. Conf. Comput. Vis.},
    year      = {2023},
    pages     = {10006-10016}
}

@inproceedings{regmi2019bridging,
  title={Bridging the domain gap for ground-to-aerial image matching},
  author={Regmi, Krishna and Shah, Mubarak},
  booktitle={Proc. IEEE Int. Conf. Comput. Vis.},
  pages={470-479},
  year={2019}
}

@InProceedings{Toker_2021_CVPR,
    author    = {Toker, Aysim and Zhou, Qunjie and Maximov, Maxim and Leal-Taixe, Laura},
    title     = {Coming Down to Earth: Satellite-to-Street View Synthesis for Geo-Localization},
    booktitle = {Proc. IEEE Conf. Comput. Vis. Pattern Recognit.},
    year      = {2021},
    pages     = {6488-6497}
}

@article{shi2019spatial,
  title={Spatial-aware feature aggregation for image based cross-view geo-localization},
  author={Shi, Yujiao and Liu, Liu and Yu, Xin and Li, Hongdong},
  journal={in Proc. Adv. Neural Inf. Process. Syst.},
  volume={32},
  pages= {10090–10100},
  year={2019}
}

@ARTICLE{10601183,
  author={Ahn, Woo-Jin and Park, So-Yeon and Pae, Dong-Sung and Choi, Hyun-Duck and Lim, Myo-Taeg},
  journal={{IEEE} Trans. Geosci. Remote Sens.}, 
  title={Bridging Viewpoints in Cross-View Geo-Localization With Siamese Vision Transformer}, 
  year={2024},
  volume={62},
  number={},
  pages={1-12},
  }

@InProceedings{Chen_2021_ICCV,
    author    = {Chen, Hao and Lagadec, Benoit and Bremond, Fran\c{c}ois},
    title     = {ICE: Inter-Instance Contrastive Encoding for Unsupervised Person Re-Identification},
    booktitle = {Proc. IEEE Int. Conf. Comput. Vis.},
    year      = {2021},
    pages     = {14960-14969}
}

@InProceedings{Xuan_2021_CVPR,
    author    = {Xuan, Shiyu and Zhang, Shiliang},
    title     = {Intra-Inter Camera Similarity for Unsupervised Person Re-Identification},
    booktitle = {Proc. IEEE Conf. Comput. Vis. Pattern Recognit. },
    year      = {2021},
    pages     = {11926-11935}
}

@ARTICLE{11010141,
  author={Li, Haoyuan and Xu, Chang and Yang, Wen and Mi, Li and Yu, Huai and Zhang, Haijian and Xia, Gui-Song},
  journal={{IEEE} Trans. Geosci. Remote Sens.}, 
  title={Unsupervised Multiview UAV Image Geolocalization via Iterative Rendering}, 
  year={2025},
  volume={63},
  number={},
  pages={1-15},
}

@article{keetha2023anyloc,
  title={Anyloc: Towards universal visual place recognition},
  author={Keetha, Nikhil and Mishra, Avneesh and Karhade, Jay and Jatavallabhula, Krishna Murthy and Scherer, Sebastian and Krishna, Madhava and Garg, Sourav},
  journal={IEEE Rob. Autom. Lett},
  volume={9},
  number={2},
  pages={1286--1293},
  year={2023},
  publisher={IEEE}
}

@inproceedings{garg2024revisit,
  title={Revisit Anything: Visual Place Recognition via Image Segment Retrieval},
  author={Garg, Kartik and Puligilla, Sai Shubodh and Kolathaya, Shishir and Krishna, Madhava and Garg, Sourav},
  booktitle={Proc. Eur. Conf. Comput. Vis.},
  pages={326--343},
  year={2024}
}

@inproceedings{shore2024bev,
  title={{BEV-CV}: Birds-Eye-View Transform for Cross-View Geo-Localisation},
  author={Shore, Tavis and Hadfield, Simon and Mendez, Oscar},
  booktitle={Proc. IEEE/RSJ Int. Conf. Intell. Robots Syst.},
  pages={11048--11055},
  year={2024}
}

@ARTICLE{10109197,
  author={Wang, Xueping and Liu, Min and Wang, Fei and Dai, Jianhua and Liu, An-An and Wang, Yaonan},
  journal={IEEE Trans. Multimedia}, 
  title={Relation-Preserving Feature Embedding for Unsupervised Person Re-Identification}, 
  year={2024},
  volume={26},
  number={},
  pages={714-723},
  doi={10.1109/TMM.2023.3270636}}

@ARTICLE{9783116,
  author={Tao, Yusheng and Zhang, Jian and Hong, Jiajing and Zhu, Yuesheng},
  journal={IEEE Trans. Multimedia}, 
  title={DREAMT: Diversity Enlarged Mutual Teaching for Unsupervised Domain Adaptive Person Re-Identification}, 
  year={2023},
  volume={25},
  number={},
  pages={4586-4597}}

@ARTICLE{10314802,
  author={Bian, Yuan and Liu, Min and Wang, Xueping and Tang, Yi and Wang, Yaonan},
  journal={IEEE Trans. Multimedia}, 
  title={Occlusion-Aware Feature Recover Model for Occluded Person Re-Identification}, 
  year={2024},
  volume={26},
  number={},
  pages={5284-5295}}

@ARTICLE{10812860,
  author={Liu, Min and Zhang, Zhu and Bian, Yuan and Wang, Xueping and Sun, Yeqing and Zhang, Baida and Wang, Yaonan},
  journal={IEEE Trans. Multimedia}, 
  title={Cross-Modality Semantic Consistency Learning for Visible-Infrared Person Re-Identification}, 
  year={2025},
  volume={27},
  number={},
  pages={568-580}}

@ARTICLE{9963608,
  author={Feng, Yujian and Yu, Jian and Chen, Feng and Ji, Yimu and Wu, Fei and Liu, Shangdon and Jing, Xiao-Yuan},
  journal={IEEE Trans. Multimedia}, 
  title={Visible-Infrared Person Re-Identification via Cross-Modality Interaction Transformer}, 
  year={2023},
  volume={25},
  number={},
  pages={7647-7659}}

@ARTICLE{10855601,
  author={Yu, Xiaoyan and Dong, Neng and Zhu, Liehuang and Peng, Hao and Tao, Dapeng},
  journal={IEEE Trans. Multimedia}, 
  title={CLIP-Driven Semantic Discovery Network for Visible-Infrared Person Re-Identification}, 
  year={2025},
  volume={27},
  number={},
  pages={4137-4150}}

@ARTICLE{9891821,
  author={Pang, Zhiqi and Zhao, Lingling and Liu, Qiuyang and Wang, Chunyu},
  journal={IEEE Trans. Multimedia}, 
  title={Camera Invariant Feature Learning for Unsupervised Person Re-Identification}, 
  year={2023},
  volume={25},
  number={},
  pages={6171-6182}}

@inproceedings{hu2018squeeze,
  title={Squeeze-and-excitation networks},
  author={Hu, Jie and Shen, Li and Sun, Gang},
  booktitle={Proc. IEEE Conf. Comput. Vis. Pattern Recognit.},
  pages={7132--7141},
  year={2018}
}

@ARTICLE{10897882,
  author={Jiang, Yan and Cheng, Xu and Yu, Hao and Liu, Xingyu and Chen, Haoyu and Zhao, Guoying},
  journal={IEEE Trans. Multimedia}, 
  title={DSAF: Dual Space Alignment Framework for Visible-Infrared Person Re-Identification}, 
  year={2025},
  volume={27},
  number={},
  pages={5591-5603}}

@ARTICLE{10962298,
  author={Zhou, Hao and Luo, Tingjin and Zhang, Jun and Liu, Liguo},
  journal={IEEE Trans. Pattern Anal. Mach. Intell.}, 
  title={Exploring the Essence of Relationships for Scene Graph Generation via Causal Features Enhancement Network}, 
  year={2025},
  volume={47},
  number={8},
  pages={6616-6630}}

@article{liu2025nonconvex,
  title={Nonconvex and discriminative transfer subspace learning for unsupervised domain adaptation},
  author={Liu, Yueying and Luo, Tingjin},
  journal={Front. Comput. Sci.},
  volume={19},
  number={2},
  pages={192307},
  year={2025},
  publisher={Springer}
}

@inproceedings{li2022cross,
  title={Cross-domain adaptive teacher for object detection},
  author={Li, Yu-Jhe and Dai, Xiaoliang and Ma, Chih-Yao and Liu, Yen-Cheng and Chen, Kan and Wu, Bichen and He, Zijian and Kitani, Kris and Vajda, Peter},
  booktitle={Proc. IEEE Conf. Comput. Vis. Pattern Recognit.},
  pages={7581--7590},
  year={2022}
}

@article{sun2021semi,
  title={Semi-supervised learning with label proportion},
  author={Sun, Ningzhao and Luo, Tingjin and Zhuge, Wenzhang and Tao, Hong and Hou, Chenping and Hu, Dewen},
  journal={IEEE Trans. Knowl. Data Eng.},
  volume={35},
  number={1},
  pages={877--890},
  year={2021},
  publisher={IEEE}
}

@article{jiang2022semi,
  title={Semi-supervised multiview feature selection with adaptive graph learning},
  author={Jiang, Bingbing and Wu, Xingyu and Zhou, Xiren and Liu, Yi and Cohn, Anthony G and Sheng, Weiguo and Chen, Huanhuan},
  journal={IEEE Trans. Neural Networks Learn. Syst.},
  volume={35},
  number={3},
  pages={3615--3629},
  year={2022},
  publisher={IEEE}
}

@inproceedings{du2021cross,
  title={Cross-domain gradient discrepancy minimization for unsupervised domain adaptation},
  author={Du, Zhekai and Li, Jingjing and Su, Hongzu and Zhu, Lei and Lu, Ke},
  booktitle={Proc. IEEE Conf. Comput. Vis. Pattern Recognit.},
  pages={3937--3946},
  year={2021}
}

@article{likas2003global,
  title={The global k-means clustering algorithm},
  author={Likas, Aristidis and Vlassis, Nikos and Verbeek, Jakob J},
  journal={Pattern Recognit.},
  volume={36},
  number={2},
  pages={451--461},
  year={2003},
  publisher={Elsevier}
}

@ARTICLE{10589921,
  author={Li, Haoyuan and Xu, Chang and Yang, Wen and Yu, Huai and Xia, Gui-Song},
  journal={IEEE Trans. Geosci. Remote Sens.}, 
  title={Learning Cross-View Visual Geo-Localization Without Ground Truth}, 
  year={2024},
  volume={62},
  number={},
  pages={1-17}
}

@article{ye2021deep,
  title={Deep learning for person re-identification: A survey and outlook},
  author={Ye, Mang and Shen, Jianbing and Lin, Gaojie and Xiang, Tao and Shao, Ling and Hoi, Steven CH},
  journal={IEEE Trans. Pattern Anal. Mach. Intell.},
  volume={44},
  number={6},
  pages={2872--2893},
  year={2021},
  publisher={IEEE}
}

@article{zhou2025cdm,
  title={Cdm-net: A framework for cross-view geo-localization with multimodal data},
  author={Zhou, Xin and Yang, Xuerong and Zhang, Yanchun},
  journal={IEEE Trans. Geosci. Remote Sens.}, 
  year={2025},
  volume={63},
  number={},
  pages={1-16},
  publisher={IEEE}
}

@article{hu2025query,
  title={Query-driven feature learning for cross-view geo-localization},
  author={Hu, Shuyu and Shi, Zelin and Jin, Tong and Liu, Yunpeng},
  journal={IEEE Trans. Geosci. Remote Sens.}, 
  year={2024},
  volume={63},
  number={},
  pages={1-15},
  publisher={IEEE}
}
\end{document}